\pdfoutput=1
\documentclass[10pt,twocolumn,letterpaper]{article}
\usepackage[T1]{fontenc}

\usepackage[pagenumbers]{wacv}      %

\usepackage{multirow}
\usepackage{colortbl}   %
\graphicspath{{figs/}}

\newcommand{\LEG}{\textsuperscript{\dag}\xspace}  %
\newcommand{\TPI}{TPI\xspace}             %
\newcommand{\ours}{PG-Lattice\xspace}     %
\newcommand{\dataset}{PickGlass-TPI\xspace}   %
\newcommand{\codeurl}{\mbox{\href{https://github.com/anoubhav/pickglass-tpi}{github.com/anoubhav/pickglass-tpi}}}
\newcommand{\avec}{\mathbf{a}}
\newcommand{\bvec}{\mathbf{b}}

\newcommand{\FNoctBlindDiff}{\ensuremath{-1.0}} %
\newcommand{\FNoctBlindDiffCI}{[\ensuremath{-2.2}, \ensuremath{-0.1}]} %
\newcommand{\BNsamePhoto}{32.0}

\newcommand{\BNrsSA}{20.5}
\newcommand{\BNrsNG}{8.9}

\newcommand{\BNrsOurs}{1.8}
\newcommand{\BNrsGpt}{14.0}
\newcommand{\BNrsEns}{9.7}
\newcommand{\BNrsSAfive}{8}
\newcommand{\BNrsNGfive}{12}
\newcommand{\BNrsOursFive}{24}
\newcommand{\BNrsGptFive}{11}
\newcommand{\BNrsEnsFive}{13}

\newcommand{\BNrsSAtwo}{12}
\newcommand{\BNrsNGtwo}{2}
\newcommand{\BNrsOursTwo}{8}
\newcommand{\BNrsGptTwo}{4}
\newcommand{\BNrsEnsTwo}{0}

\newcommand{\BNhumNG}{37.3}

\newcommand{\BNgptNG}{27.1}
\newcommand{\BNgptSA}{28.8}
\newcommand{\BNensNG}{25.4}

\newcommand{\BNoursNG}{34.8}
\newcommand{\BNoursSA}{41.7}
\newcommand{\BNgptMinusHumNG}{\ensuremath{-10.2}}
\newcommand{\BNgptMinusHumNGci}{[\ensuremath{-17.9}, \ensuremath{-2.7}]}

\newcommand{\BNcvSA}{0.19}
\newcommand{\BNcvNG}{0.16}

\newcommand{\BNlogGptMinusHum}{\ensuremath{-0.07}}
\newcommand{\BNlogGptMinusHumci}{[\ensuremath{-0.16}, \ensuremath{+0.02}]}

\newcommand{\BNrfourN}{64}

\newcommand{\BNrfourNGSA}{15}
\newcommand{\BNrfourNGours}{28}

\newcommand{\BOlinesBigOct}{48} %

\newcommand{\BToursOctLinesNG}{36}

\definecolor{wacvblue}{rgb}{0.21,0.49,0.74}
\usepackage[pagebackref,breaklinks,colorlinks,allcolors=wacvblue]{hyperref}

\def\wacvPaperID{*****} %
\def\confName{WACV}
\def\confYear{2027}

\title{Image Quality and the Reliability of Fabric-Density Measurement:\\
Counting Threads from Smartphone Photos Through a Pick Glass}

\author{Anoubhav Agarwaal\\
Indian Institute of Technology Madras\\
{\tt\small anoubhav.agarwaal@gmail.com}
}

\begin{document}
\twocolumn[{%
\renewcommand\twocolumn[1][]{#1}%
\maketitle
\begin{center}
  \vspace{-0.6em}
  \includegraphics[width=\textwidth]{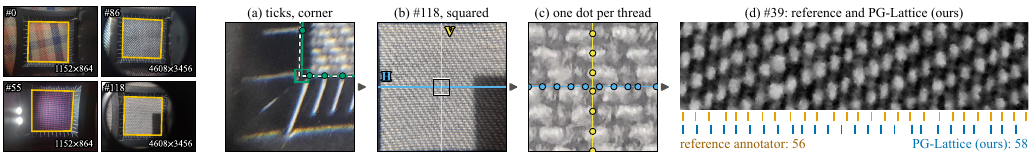}
  \captionof{figure}{\dataset and the task: from a phone photo taken through a pick glass (a 10$\times$ magnifier), count the threads that cross its one-inch window in each direction (threads per inch). Left: factory photos differ in glare, shadow, color and resolution (yellow: hand-marked window; numbers: image size in pixels). (a) On photo \#118, ticks engraved on the glass locate the window's corner (green: fit to the ticks; dashed: hand-marked edges). (b) The window, warped to a square, with the horizontal (H) and vertical (V) counting lines. (c) The box in (b), enlarged: \ours, our training-free counter, puts one dot on each thread crossing a line. (d) On part of the H line of photo \#39, the textile-trained reference annotator and \ours mark nearly the same threads (numbers: counts over the whole line); a non-expert annotator, our human baseline, marks 111, about twice as many, and the photo alone cannot settle which is right (\cref{sec:exp_octave}).}
  \label{fig:dataset}
\end{center}
\vspace{0.4em}
}]
\begin{abstract}
Thread density, the threads per inch along and across a woven fabric, is a basic specification that buyers check whenever fabric is sold.
Without laboratory instruments, inspectors count it slowly, by eye, through a pick glass, a folding 10$\times$ magnifier with a one-inch window.
A smartphone photo through that glass could make the check faster and leave a record.
We judge such a photo by what it is used for, as in task-based image quality: it is good if the thread count from it is within $\pm$5\% of the true count, a common tolerance in fabric specifications.
We introduce \dataset: 76 factory scenes of 44 fabric patches, each counted from scratch in an open annotation tool by a textile-trained annotator, whose counts are the reference.
On it we benchmark frontier and open vision-language models asked directly for the counts, CNNs, counting specialists and a training-free spectral counter (\ours).
No method agrees with the reference within the tolerance.
The best methods are off by about a quarter: a small open model that we tuned only on synthetic images (Molmo2-4B, 25\%) is level with the best frontier model, GPT-6.1-sol (27\% when asked directly, 25\% given four magnified close-ups of the window).
Both err less than a non-expert annotator who counted the same photos (37\%), but only the tuned model also beats, on both error measures, a trivial baseline that guesses each scene's density as the average of the other scenes.
\ours, light enough to run offline, most often repeats its count within 5\% across two photos of the same patch, but its misses, like GPT-6.1-sol's errors, are often a factor of two.
Even the reference repeats a count across such re-shoots only to a median 9\%, so one photo cannot yet be shown to meet $\pm$5\%; the image quality that matters is reliability across re-shoots.
On these counts, generic no-reference quality scores chosen before labeling do not rank \ours's error, and generative enhancement raises perceptual scores but does not reliably improve the count.
Code, annotation tool and all predictions are at \codeurl; photos and counts will follow after review.
\end{abstract}

\section{Introduction}
\label{sec:intro}

Thread density, the number of warp \emph{ends} and weft \emph{picks} per inch (\TPI), is a basic entry of woven-fabric specifications that buyers check whenever fabric is sold~\cite{iso2024threads,astm2017d3775}.
A standard hand tool for this count in many countries is the pick glass (also called a linen tester or counting glass): an inspector places this folding 10$\times$ magnifier, whose base has a nominal 1\,inch$\times$1\,inch opening, on the cloth and counts yarns with a needle at five or more places per piece~\cite{bis1981is1963,iso2024threads}.
Counting at 10$\times$ is slow, and it matters most where laboratories are out of reach: in India, for example, most woven fabric comes from powerloom units and from handloom households, a sector that is mostly rural and employs mostly women~\cite{ibef2026textiles,ibef2026powerloom,dchl2024handloombrief}, while about 85\% of Indian households already own a smartphone~\cite{mospi2025telecom}.

Image-based density measurement has a 30-year history~\cite{xu1996fft,meng2022review}, but almost all of it assumes a controlled capture (scanners, backlights, microscopes, loom cameras; \cref{sec:related}), none of which is on the inspection table of a small mill; a smartphone is.
We therefore keep the inspector's own tool and photograph the fabric \emph{through the pick glass}.
The glass's opening and the scales engraved on its frame are a metric fiducial in every photo: the ticks locate the window, and counting yarns across the rectified window gives threads per window side, which is \TPI if the opening is one inch (\cref{fig:dataset}).

This setup turns a textile measurement into an image-quality problem: photos show LED hot spots, the phone's shadow, defocus at 10$\times$, hand shake, tilt and a fourfold range of resolution.
As in task-based quality for biometrics and medical imaging~\cite{grother2007quality,saeed2022taskamenability,wang2025miqa}, we call a pick-glass photo good if the threads counted from it fall within the $\pm$5\% tolerance of product standards.
We ask how close today's methods come to that tolerance and whether image quality can be judged before counting.
To find out, we collect \dataset on a working factory floor, build an open annotation tool in which a textile-trained annotator, our reference, and a non-expert, a human baseline, count every photo from scratch, and test a wide range of methods on the same photos, from frontier vision-language models (VLMs) asked directly for the counts to models small or light enough to run on site: CNNs, counting specialists, a small open VLM that we tune on synthetic crops, and a training-free spectral counter (\ours).
One caveat bounds every result: the reference repeats a count across two photos of the same patch (re-shoots) only to a median 9\%, so one photo count cannot yet be shown to meet $\pm$5\%.

\paragraph{Contributions.}
\begin{enumerate}
\item \textbf{Task, data and tools.} We define image quality for pick-glass thread counting by the $\pm$5\% tolerance and collect \dataset: 76 factory scenes of 44 fabric patches, each counted from scratch by both annotators in an open annotation tool (\cref{fig:tool}), with patches photographed more than once so that one can test whether a count repeats; and a ruler that finds the counting window from the glass's engraved scales in all but one scene (\cref{sec:data,sec:ruler}). Code, tool and predictions: \codeurl.
\item \textbf{From frontier models to on-site counters.} We score frontier and open VLMs, CNNs, counting specialists and training-free counters against the textile-trained reference, and tune a small open VLM on synthetic crops alone: it is off by about a quarter, level with the best frontier VLM and below a non-expert annotator, and unlike that VLM beats a trivial baseline on both error measures; much of the remaining error is a factor of two; and the training-free counter most often repeats a count within 5\% across re-shoots, but its misses are often twofold (\cref{sec:cnn,sec:exp_main,sec:exp_octave,sec:exp_repeat}).
\item \textbf{What image quality does and does not tell.} Generic quality scores chosen before labeling neither rank \ours's error nor flag where the non-expert and the reference disagree; generative enhancement raises perceptual scores without reliably improving the count; and counts break with too few pixels per yarn, which up-sampling does not restore (\cref{sec:exp_iqa,sec:exp_quality,sec:exp_genai}).
\end{enumerate}

\section{Related Work}
\label{sec:related}

\paragraph{Fabric density from images.}
Image-based methods measure fabric density with Fourier spectra~\cite{xu1996fft,ravandi1995fourier,yildiz2014density,zhang2015transmittance}, wavelets~\cite{jing2015wavelet,zhao2018wavelet}, co-occurrence~\cite{lin2002cooccurrence}, deskewed gray-level projection~\cite{pan2010hough,jing2014skew,zhang2014subimages,xiang2022yarndyed,karmuse2025robust} and CNNs that localize yarns~\cite{meng2019msnet,meng2021mtmsnet,ding2024mtfnet,tan2024segmentation} (review:~\cite{meng2022review}).
These systems control distance, focus, illumination, flatness or scale (scanners, transmitted light~\cite{zhang2017backlight}, loom and sewing-zone cameras~\cite{schneider2015onloom,hu2018threadcounting}, an enclosed 64\,MP camera box~\cite{alipio2026threadcounter}).
The closest works put a phone on a printed backlight with a calibration square~\cite{xiang2018portable}, or behind a loupe to estimate yarn fineness (Ne count)~\cite{rofidatunnissa2025yarncount}.
A commercial AI pick counter with a macro camera has been announced without method or accuracy data~\cite{paramount2026digipick}, and the only public density-labeled set, FabricNET~\cite{seckin2023fabricnet,seckin2025fabricnethf}, has 130 handheld-microscope images without a scale reference.
To our knowledge, no prior work estimates thread density from phone photos taken through the inspector's own pick glass; we do, with the glass as the only aid besides the phone, and score every count against human counts and the $\pm$5\% tolerance.

\paragraph{Counting with foundation models and class-agnostic counters.}
Vision-language models (VLMs) count worse as counts grow~\cite{paiss2023countclip,qharabagh2024lvlmcount,tamarapalli2025countqa,rong2026unicbench}; four frontier VLMs averaged 47\% on counting the rows and columns of synthetic grids~\cite{rahmanzadehgervi2024blind}, a clean abstraction of our task.
Point-then-count models~\cite{deitke2025molmo,clark2026molmo2} and exemplar-based counters~\cite{ranjan2021famnet,liu2022countr,dukic2023loca,pelhan2024geco,amininaieni2024countgd} target tens to hundreds of separate objects, whereas a 1-inch window holds thousands of touching crossings.
VLMs have been tested on fabric properties~\cite{wang2025mllmfabric}, fabric structure~\cite{wang2026taha} and crochet~\cite{li2025crochetbench}, but to our knowledge not on thread counting.
Reading half or double the true number of periods is the classic octave error of pitch estimation~\cite{decheveigne2002yin}; period ambiguity also limits repetition counting~\cite{dwibedi2020repnet} and lattice detection~\cite{hays2006lattice,park2009lattice}.
Like Mueez and Vyas~\cite{mueez2026grain}, who compare foundation-model, CNN and VLM methods on ASTM E112 grain size, we compare model families on a standards-based measurement; unlike prior counting work, we score them against a stated tolerance, tune an open VLM on exact synthetic points, and trace much of their error to a factor-of-two (octave) ambiguity.

\paragraph{Capture quality, camera simulation and enhancement.}
Blur, noise, compression and resolution degrade recognition~\cite{dodge2016quality,vasiljevic2016blur,hendrycks2019imagenetc}.
Perceptual image-quality assessment (IQA)~\cite{mittal2012brisque,fang2020spaq,wu2024qalign,ke2021musiq} is trained to predict human opinion scores.
Task-based image quality is defined by performance on a task~\cite{barrett1990objective,barrett2004foundations} and is standardized in biometrics as sample utility~\cite{iso2024biometricquality}; it is used in face recognition~\cite{grother2007quality,terhorst2020serfiq}, medical imaging~\cite{saeed2022taskamenability}, machine-centric IQA~\cite{wang2025miqa} and document capture~\cite{kumar2014docquality}, and to decide when to abstain~\cite{geifman2017selective}.
WVAQ precedents report large losses along blur and resolution sweeps~\cite{bhatta2024blur,pangelinan2026probegallery}, enhancement whose perceptual scores do not predict detection gains~\cite{aryashad2026defogging,saleem2026enhancement}, effective resolution as distinct from generic IQA~\cite{kansy2023effres}, and quality gates that cut review cost~\cite{giroinieto2026costsavings} or reject uncertain grades~\cite{roy2026gemstone}.
Unlike these and camera-simulation and re-capture benchmarks for recognition~\cite{brooks2019unprocessing,carlson2018camera,baek2024imagenetes,kim2026imagenetses}, whose label does not change with the capture, we define a photo's quality by whether its count meets the tolerance, check that this count survives each simulated capture stage, and test whether generic quality scores, resolution and generative restoration~\cite{wang2021realesrgan,yue2025invsr,wang2026seedvr2}, which can invent texture, predict or change it.

\section{Task and the \dataset Dataset}
\label{sec:data}

\paragraph{Task.}
Given one photo taken through the pick glass, estimate (i) the window's four corners; (ii) $h$ and $v$, the numbers of vertical and horizontal yarns crossing a horizontal and a vertical line across the window, counting the threads actually present~\cite{iso2024threads}; (iii) the weave, scored separately, and the warp direction, which maps $h,v$ to ends and picks but is not scored, as a window rarely shows a selvedge (the woven edge); and (iv) a confidence that allows abstention.
A homography from the corners to an $S{\times}S$ crop~\cite{hartley2004mvg} makes the image metric, so the crossing counts are threads per window side at any resolution; they are \TPI only if the opening is one inch at the fabric plane, which the glass's own scales leave uncertain by a few percent (Sec.~S9); as the annotators and methods count the same rectified window, this scale factor cancels in every reported error, and absolute \TPI needs one factor per glass.
Standards average five or more counts over 1--10\,cm~\cite{bis1981is1963,iso2024threads}, and we found no precision statement beyond a four-fabric ASTM study~\cite{astm1984rrd131067}; tolerances are left to product specifications, \eg $\pm$5\%, $+5/{-2.5}$\%, or $-5$\% on the mean of the readings~\cite{bis2009is179,bis2009is15853,bis2003is745}, so one window is a screening measurement whose target accuracy is $\pm$5\%.

\paragraph{Capture and provenance.}
The photos were taken on a working factory floor through a commodity 10$\times$ pick glass pressed onto the fabric, with uncontrolled distance, tilt, focus and lighting; no file has EXIF, and every JPEG was re-saved at quality 75.
An audit of the 150 files finds 76 distinct \emph{scenes}, which show 44 fabric patches; the other files are lower-resolution copies, duplicates, PNG decodes and synthesized crops, and the synthesized crops are never scored against human counts (Sec.~S11).
Fifty scenes have a full-resolution photo (window about 1900\,px wide) and 26 only a quarter-width copy (about 500\,px).
Statistics use one row per scene unless stated, and every split is grouped by scene.
The author took the photos with the mill's agreement. We will release them with their scene map, polygons, rectified crops, human counts with dot positions and timing logs, the annotation tool, all predictions and the code, with a datasheet~\cite{gebru2021datasheets}.

\begin{figure}[t]
  \centering
  \includegraphics[width=\linewidth]{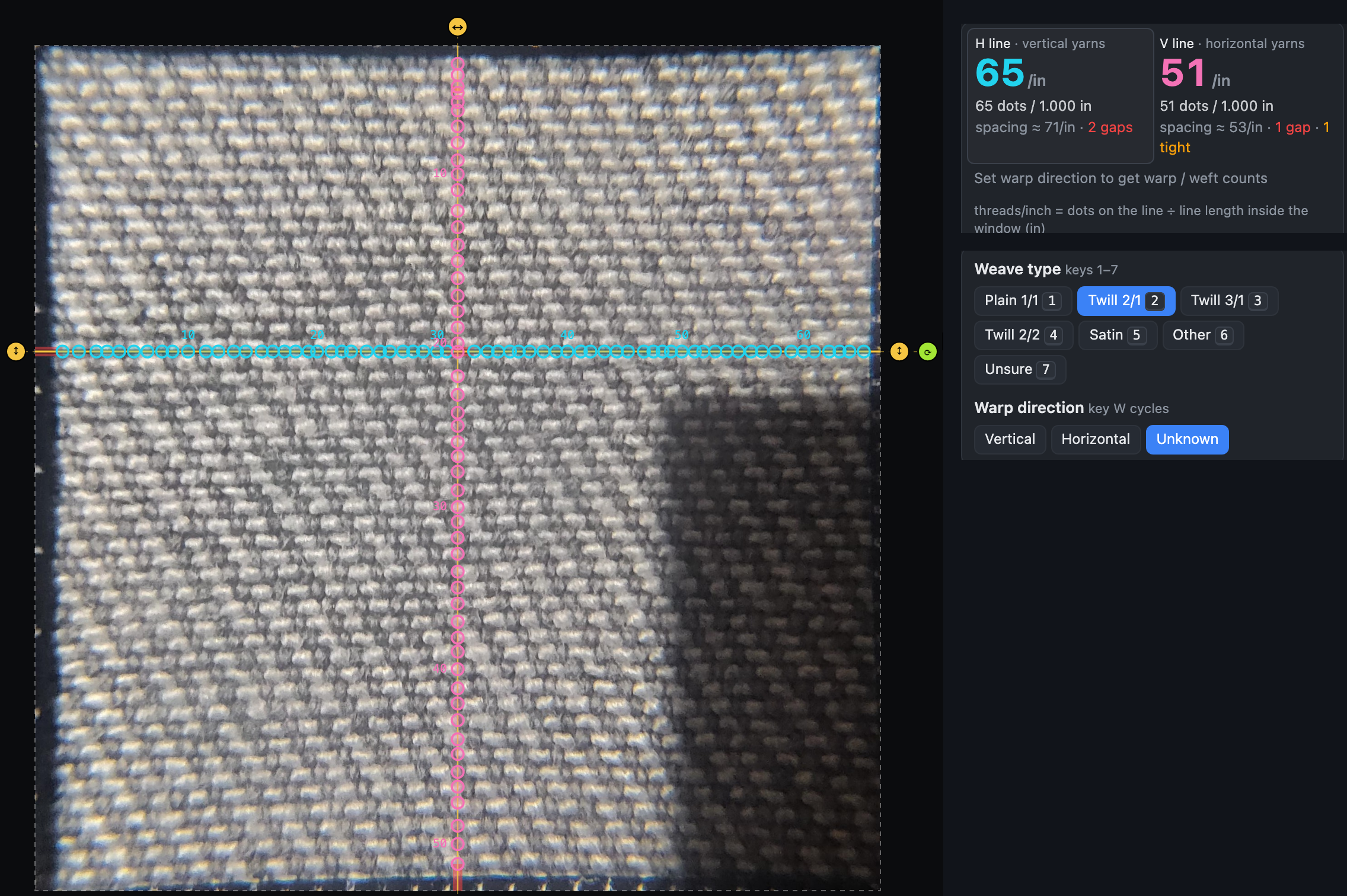}
  \caption{The open annotation tool on scene \#118, with the reference annotator's dots. One dot marks each yarn crossing the horizontal (cyan) or vertical (pink) guide line; the side panel shows only values computed from the annotator's own dots (count, spacing, gaps), never a machine count, and records the weave. The tool and a demo set are at \codeurl.}
  \label{fig:tool}
\end{figure}

\paragraph{Scene-aware thread-count annotation.}
In an open browser tool (\cref{fig:tool}) the annotator places one dot per yarn crossing each of two guide lines on the rectified crop ($h$, $v$ = dots per inch of line) and records the weave, warp direction (always left unknown), capture-quality flags and a confidence; only one representative photo per scene is annotated, and copies inherit its counts (Sec.~S11).
No machine count or weave guess is shown, and machine dots appear only in four guide examples; the tool accepts a photo only when both lines are counted across the window or an exception is noted, and on 38 of the 76 scenes it showed a ``check carefully'' or ``take care'' note when the spectral estimators (\cref{sec:ours}) disagreed, with the reason but no count (Sec.~S3).
Two paid external annotators, hired after both matched the legacy hand counts (below) closely in a public screening test, each counted every scene from scratch: one with textile training, whose counts are the reference for every result, and one without, a human baseline for how far a careful non-expert gets.
The textile-trained annotator took more than twice as long per photo (Tab.~S2) and repeated a count across re-shoots more closely (\cref{tab:repeat}).
On the same photo the two counts differ by a median \BNsamePhoto\% of their mean, and by more than a fifth even on the scenes without the note (Tab.~S2): above the 10\% beyond which the plan fixed in advance calls a per-photo human count too noisy to test $\pm$5\%. %

\paragraph{Re-shoots and the reference.}
Label-free image matching (Sec.~S11) finds 41 pairs of scenes that photograph the same fabric patch again (24 patches), so each pair's true counts agree up to an $h$/$v$ swap: they measure how well the annotators and every method repeat a count without ground truth (\cref{sec:exp_repeat}).
\Cref{tab:main150} scores against the textile-trained annotator, whose counts were also the blind test set, opened only after every method was frozen. Scoring against each annotator alone was fixed before the reference counts were opened (Tab.~S7); making the textile-trained counts the main reference, as they repeat more closely and the mean of both matches neither reading where the two differ twofold, was decided after both annotators' counts were seen and departs from the plan, which scored against that mean (Tab.~S6; \cref{sec:exp_main}).

\paragraph{Legacy hand counts.}
We also use 13 hand-counted window crops of plain and 2/1 twill fabrics from an earlier project (the photos are lost); their labels are noisy (Sec.~S2).
No counter was fitted on these 13 crops except two post-hoc variants, and the set informed some design choices (Sec.~S1); every method is scored \emph{axis-matched} ($h\leftrightarrow$ warp) on the same crop, trimmed and resized to a square as the VLMs receive it.

\section{Methods}
\label{sec:methods}

All counters receive the same rectified 1-inch crop (from the annotated corners unless stated otherwise) and return $(h, v)$, a weave label and, where available, a confidence.

\subsection{Window from the glass's ruler}
\label{sec:seg}
\paragraph{The glass's own ruler.}
\label{sec:ruler}
The scales engraved on the frame are a calibration target in every photo: we detect the ticks, index them on a 1-D projective lattice per row and fit a robust homography to all tick feet, measuring the window side per photo; it was developed on these photos without counts, and the polygons only validate it (Sec.~S9).

\subsection{A training-free spectral counter (\ours)}
\label{sec:ours}
As a transparent offline baseline whose spectrum can be inspected, we use a training-free counter that reads the yarn lattice from the 2-D power spectrum; it also explains the half-or-double readings (Sec.~S16) and supplies the resolvability check below.

\paragraph{Pre-processing.}
Color is projected onto the first principal axis of the high-passed RGB vectors, which keeps the dominant color contrast even when it is chroma only; specular glare is masked; and local contrast normalization removes LED hot spots and shadow gradients (Sec.~S4). Processing runs on at most $1024^2$\,px (side $N$), and counts are cycles per crop side, so they are per window side at any resolution.

\paragraph{Why a lattice.}
A woven fabric is a 2-D lattice: the vertical yarns give a reciprocal vector $\avec$ near the $u$ axis of the spectrum, the horizontal yarns give $\bvec$ near the $v$ axis, and the interlacing adds a sub-lattice $W$: $(\tfrac12,\tfrac12)$ for plain weave, $\tfrac1R(1,\pm1)$ for a twill with repeat $R$, and $(\mp\tfrac25,\tfrac15),(\tfrac15,\pm\tfrac25)$ for a 5-end satin.
For regular weaves only multiples of the yarn frequencies lie on the axes, so weave repeats never add spurious on-axis peaks; color stripes and checks can, but they are coarser than the yarns.
Classical FFT practice uses this implicitly~\cite{xu1996fft}; we score whole lattice hypotheses (Sec.~S4), as in lattice detection for periodic texture~\cite{hays2006lattice,park2009lattice}.

\paragraph{Hypothesis search.}
The Hann-windowed log power spectrum is whitened by its radial median into robust $z$-scores $g(\mathbf{p})$.
Hypotheses $H=(\avec,\bvec,W)$ combine the 10 strongest near-axis peaks per family (within $\pm15^\circ$) with eight sub-lattices $W$ (none, plain, twill $R\in\{3,4\}$ in Z and S direction, 5-end satin in both orientations).
The score of $H$ (Eq.~S1 in the supplement) adds the evidence at the points of the lattice and $W$, and subtracts the evidence at on-axis sub-harmonics, at finer peaks that $H$ leaves unexplained (as yarns are the finest periodic structure), at harmonics stronger than their fundamental, and a rotation prior.
The winning $\avec,\bvec$ are refined to the centroid of their peak clusters, and $h=|a_u|$, $v=|b_v|$.

\paragraph{Fusion, confidence and resolvability.}
Two independent estimators, the period of a deskewed projection profile found with YIN~\cite{decheveigne2002yin} and the median FFT estimate of overlapping tiles, can override the lattice when they agree with each other and disagree with it; a confidence is fitted on synthetic data only, and the weave label comes from the winning sub-lattice (design ablation, all on synthetic images: see Sec.~S4).
A second label-free check, \emph{resolvability}, tests whether the photo is sharp enough to show yarns as fine as claimed: with $s$ the Laplacian-to-image variance ratio and $f=\max(h,v)/N$ the finer claimed frequency in cycles per working pixel, $\eta=s/(4\sin^2\pi f)^2$ is about 1 for a sharp grating at $f$ and far below 1 when the counter has locked onto blur or interpolation artifacts.
We also report simpler estimators, including the maximum near-axis FFT peak and a log-Gabor bank~\cite{field1987loggabor}.
A combined counter, pre-registered before any human label, takes per axis the geometric mean of \ours and the log-Gabor bank where the two nearly agree, and \ours elsewhere; it helps on fresh synthetic crops but barely changes the error on \dataset (\cref{tab:main150}; Sec.~S17).

\subsection{Vision-language models}
\label{sec:vlm}
We query five closed frontier VLMs and two open-weight ones a mill could host~\cite{qwen2026qwen38,gemma2026gemma4} on the same crop resized to 1536\,px (up-sampled about threefold on the quarter-width scenes; on scenes with both widths, width changes the error by under 4 points, post hoc; Sec.~S6) at temperature 0 and the lowest reasoning setting (models: \cref{tab:main150} and Sec.~S6).
Unless stated, a VLM receives the direct prompt P1, which asks for $h$, $v$ and the weave as JSON (other prompts: Sec.~S6; model ids, costs and raw outputs: code release).

\subsection{Learned and specialist counters}
\label{sec:cnn}
\paragraph{CNNs and counting specialists.}
An ImageNet-pretrained~\cite{deng2009imagenet} residual CNN~\cite{he2016resnet} (Sec.~S5) regresses the log number of yarns crossing a 224-px tile per axis, plus the weave, from 4{,}800 analytically rendered crops with a simulated phone capture; the estimate is the median over tiles, three scales and flips.
Variants add label-exact augmentation, fine-tune on \ours's counts of the scenes it is most confident on (Sec.~S15), also with a second backbone~\cite{tan2019efficientnet}, or fine-tune (FT) on the non-expert's counts with five-fold cross-validation grouped by patch, so the reference stays their blind test (Sec.~S1); a CNN--VLM ensemble combines the patch-grouped CNN with GPT-6.1-sol P1 by a per-axis geometric mean (\cref{sec:exp_main}).
Training its CNN also on 843 augmented copies did not help ($+1.0$, CI [$-1.2$, $+2.8$]; Sec.~S1); copies repeat their scene's count, so they are not test data.
A point-then-count VLM, Molmo2-8B~\cite{clark2026molmo2}, and an exemplar-based counter, CountGD~\cite{amininaieni2024countgd}, serve as counting specialists (Sec.~S7).
\paragraph{A small VLM tuned on synthetic crops (ours).}
We tune Molmo2~\cite{clark2026molmo2}, an open point-then-count VLM, because pointing lets synthetic crops supervise it yarn by yarn, and its 4B version is small enough to LoRA-tune~\cite{hu2022lora} on two consumer GPUs (5.9 GPU-hours) and for a mill to host; untuned, it is off by 83\% on the legacy crops (Sec.~S7).
Our crop generator knows exactly where each yarn crosses a guide line, through rectification, resampling and perspective; on 12{,}000 such crops we tune Molmo2-4B to point at the crossings and to answer P1 (\cref{sec:vlm}).
No real image or label is used, and checkpoint and readout were fixed on synthetic validation in one run, before any human count was scored (Sec.~S7).

\section{Experiments}
\label{sec:exp}

\paragraph{Protocol and metrics.}
We report the mean absolute percentage error (MAPE), the shares of the $2n$ counts ($h$ and $v$ of $n$ images) within 5\% and 10\%, and, on \dataset, the number of scenes with both counts within 5\%, fixed in the scorer before labeling (MAE and weave agreement, against the mean of both annotators: Tab.~S6).
Against the reference we also report the symmetric error $|\log_2(\text{count}/\text{reference})|$, fixed in advance, which charges a halved and a doubled count alike, whereas MAPE charges them 50\% and 100\%.
The unit is the scene, scored on the crop the methods see (per-file scores, against the mean: Sec.~S17).
95\% confidence intervals (CIs) are percentile bootstraps over images or scenes~\cite{efron1993bootstrap} (patch-clustered CIs of the key differences: Sec.~S3); paired differences resample jointly and use unrounded values, with Wilcoxon signed-rank tests for MAPE and exact McNemar tests for shares; unanswered queries are not imputed.
Status tags: \emph{pre-registered} (analysis plan frozen, with a hash, before any label), \emph{blind} (method fixed before the reference counts were opened, then scored against them), \emph{post hoc} ($^\ddagger$; chosen after seeing some counts), \emph{exploratory} (analyzed but not a planned endpoint); \LEG marks the 13 legacy crops (26 counts).

\subsection{How well do methods count?}
\label{sec:exp_main}

\begin{table}[t]
\centering
\footnotesize
\setlength{\tabcolsep}{2.8pt}
\resizebox{\ifdim\width>\linewidth\linewidth\else\width\fi}{!}{%
\begin{tabular}{@{}lcccc@{}}
\toprule
 & MAPE & \multicolumn{2}{c}{\% of counts} & Both \\
\cmidrule(lr){3-4}
Method & [95\% CI] & $\le$5\% & $\le$10\% & $\le$5\% \\
\midrule
\csname @@input\endcsname tables/main150_rows_refB.tex
\bottomrule
\end{tabular}}
\caption{Thread counts on \dataset against the textile-trained reference annotator (76 scenes). First row: the non-expert's counts of the same photos; leave-one-out mean: each scene guessed as the mean of the other scenes' reference counts. MAPE in \%; $\le$5\%, $\le$10\%: share of counts within 5\%, 10\%; both $\le$5\%: scenes with both counts within 5\%; bold: best method per count column; shaded: our methods. Cascade: \cref{sec:exp_iqa}. $^\S$Four magnified insets of the window (pre-declared, Sec.~S6). $^{k/76}$: scenes answered; $^\ast$scored only on the scenes it was not trained on; $^\ddagger$post hoc (trained on, or chosen after, the non-expert's counts; CountGD: after the legacy results). Unmarked methods were fixed before any human count of \dataset, all before the reference counts were opened; Molmo2-4B, insets and $^\ddagger$ rows joined after labeling. Against the mean of both annotators (also MAE, weave) and the non-expert: Tabs.~S6, S7.}
\label{tab:main150}
\end{table}

No method agrees with the textile-trained reference within the tolerance (\cref{tab:main150}); paired tests are post hoc unless fixed in advance.
Of blind methods, the lowest errors come from our tuned Molmo2-4B and GPT-6.1-sol given four magnified insets of the window (\cref{tab:main150}), level on both errors; the latter beats the mean-density guess on MAPE only (post hoc). Molmo2-4B is also level on MAPE with GPT-6.1-sol asked directly (P1; $-1.9$ points, CI [$-6.5$, $+2.6$]) and closer to the reference on the symmetric error ($-0.13$, CI [$-0.22$, $-0.05$]).
Both beat the non-expert on MAPE (GPT-6.1-sol P1: \BNgptMinusHumNG{}, CI \BNgptMinusHumNGci; Molmo2-4B: $-12.1$, CI [$-19.7$, $-4.6$]).
On the symmetric error Molmo2-4B stays closer ($-0.20$, CI [$-0.30$, $-0.11$]); GPT-6.1-sol P1 ties with the non-expert (\BNlogGptMinusHum, CI \BNlogGptMinusHumci).
MAPE charges the non-expert's twofold misses, mostly doublings, more than GPT-6.1-sol P1's halvings (\cref{sec:exp_octave}).
Molmo2-4B beats guessing the other scenes' mean density (8.0, CI [2.9, 13.3]), also over patches and on the symmetric error; GPT-6.1-sol P1's margin (6.1, CI [0.6, 11.7]) holds on neither (Sec.~S3).
Other frontier VLMs err more; general open-weight ones fail.
The training-free counters do not beat the mean-density guess (\ours: $+1.6$, CI [$-6.1$, $+9.4$]), mostly as they read half or double (\cref{sec:exp_octave}); yet they lead on counts within 5\%, as many errors are small or about twofold.
The CNN--VLM ensemble ($^\ddagger$; GPT-6.1-sol plus a CNN trained on the non-expert's counts) does not reliably beat GPT-6.1-sol P1 in its blind test fixed in advance ($-1.7$, CI [$-4.2$, $+0.7$]).
Some rankings depend on the reference: against both annotators' pre-registered mean (Tab.~S6), GPT-6.1-sol P1 ties with the mean-density guess ($-2.1$, CI [$-6.9$, $+2.6$]) and Molmo2-4B's MAPE is 28.2\%, against 28.7\% for that guess; against the non-expert, Molmo2-4B's error rises to 47.6\% (Sec.~S7), \ours's to \BNoursSA\%, GPT-6.1-sol P1's only to \BNgptSA\% (Tab.~S7).
Weave is not interpreted: the annotators' weave labels agree no better than chance (Tab.~S2), so the reference is for counts only (weave scores: Tab.~S6).

\subsection{Half or double}
\label{sec:exp_octave}

\begin{figure}[t]
  \centering
  \includegraphics[width=\linewidth]{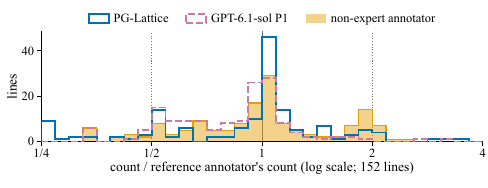}
  \caption{Methods and the non-expert annotator often read half or double the textile-trained reference (post hoc for the methods): each count of \ours, GPT-6.1-sol P1 and the non-expert over the reference's count on the same line (below 1/4: first bin).}
  \label{fig:octave}
\end{figure}

Much of the remaining error is a factor of two (\cref{fig:octave}); a count reads half or double when $|\log_2(\text{ratio})|$ lies within 0.3 of 1.
Against the reference, \ours reads half or double on \BToursOctLinesNG{} of the 152 lines (two-thirds of them halvings), GPT-6.1-sol P1 on 37 (almost all halvings) and the non-expert on \BOlinesBigOct{} (mostly doublings).
Choosing \ours's octave with the reference counts, a post-hoc upper bound that also corrects threefold and larger misses, would cut its MAPE (\cref{tab:main150}) to 10.8\%.
Before the reference counts were opened, we fixed a rule for the lines where \ours reads half or double the non-expert (\BNrfourN{} on this band, the same outcome on every band; Sec.~S16): if the reference agreed (within 10\%) with the non-expert on at least 80\% of them, the machines misread; if with \ours on at least half, the human reading is ambiguous; otherwise both readings stand.
It agrees with the non-expert on \BNrfourNGSA{} and with \ours on \BNrfourNGours{}, so both readings stand: from the photo alone neither count can be called right, and on some lines the reference itself may be on the other one.
The twofold readings look partly like a choice of what counts as one yarn: both annotators place fairly evenly spaced dots (median coefficient of variation of the spacing \BNcvNG{} for the reference and \BNcvSA{} for the non-expert), so neither counted yarn edges twice, though the non-expert also switches readings between photos of the same patch (\cref{tab:repeat}).
At 10$\times$ a single yarn can show its twisted strands (plies), two neighboring yarns can work as a pair, and a float (a stretch of yarn passing over two crossing yarns) covers both, so one reader can see one yarn where another sees two (Fig.~\ref{fig:dataset}(d)).
Two blind octave checks do not settle it: doubling a count only when the gray profile shows a second yarn changes \ours's error by \FNoctBlindDiff{} points (CI \FNoctBlindDiffCI), and a learned selector does not reliably lower it (Sec.~S16).

\subsection{Repeatability across re-shoots}
\label{sec:exp_repeat}

\begin{table}[t]
\centering
\footnotesize
\setlength{\tabcolsep}{2.4pt}
\begin{tabular}{@{}lcccc@{}}
\toprule
 & \multicolumn{3}{c}{Re-shoots (41 pairs)} & MAPE (\%) \\
Counts from & median & $\le$5\% & $\approx$2$\times$ & against \\
 & (\%) & (pairs) & (pairs) & reference \\
\midrule
Reference (textile-trained) & \BNrsNG & \BNrsNGfive & \BNrsNGtwo & -- \\
Non-expert annotator & \BNrsSA & \BNrsSAfive & \BNrsSAtwo & \BNhumNG \\
\midrule
\rowcolor{black!6}\ours (ours) & \textbf{\BNrsOurs} & \textbf{\BNrsOursFive} & \BNrsOursTwo & \BNoursNG \\
\rowcolor{black!6}Molmo2-4B + LoRA (ours) & 14.3 & 11 & 1 & 25.2 \\
GPT-6.1-sol P1 & \BNrsGpt & \BNrsGptFive & \BNrsGptTwo & \BNgptNG \\
CNN--VLM ensemble$^\ddagger$ & \BNrsEns & \BNrsEnsFive & \textbf{\BNrsEnsTwo} & \BNensNG \\
\bottomrule
\end{tabular}
\caption{\ours repeats a count within 5\% on the most pairs, but reads more pairs about twofold apart than the other methods (our tuned Molmo2-4B: one). Left, label-free: pairs of photos of one patch; per pair the larger relative axis difference (better of direct and swapped $h,v$): median, and pairs with both axes within 5\% or with an axis about twofold apart. Right: MAPE against the reference (\cref{tab:main150}). Bold: best per re-shoot column; shaded: our methods. $^\ddagger$Trained on the non-expert's counts, out of fold by patch; the reference is its blind test; the non-expert's and Molmo2-4B's re-shoot rows are post hoc.}
\label{tab:repeat}
\end{table}

Re-shoots test, without ground truth, whether a count repeats (\cref{sec:data}): even the textile-trained reference repeats a count only to a median \BNrsNG\% (\cref{tab:repeat}), so one photo's count cannot be shown to meet $\pm$5\%, while the non-expert reads the most pairs about twofold apart.
\ours repeats a count within 5\% on the most pairs, but reads more pairs about twofold apart than the other methods, and a repeated count need not be right: of the \BNrsOursFive{} pairs on which it repeats within 5\%, on 8 both photos read about half or double the reference on an axis (post hoc).
Our tuned Molmo2-4B repeats less tightly, like GPT-6.1-sol P1, but reads only one pair twofold apart (post hoc).

\subsection{Generic quality scores do not rank the error}
\label{sec:exp_iqa}

Scored against the reference (post hoc; the test frozen before any human count, against the mean of both, gives the same outcome, Sec.~S8), generic no-reference scores do not rank \ours's per-scene error (Spearman $\rho$; scores signed so that higher means better, so a useful score has $\rho<0$): BRISQUE~\cite{mittal2012brisque} and NIQE~\cite{mittal2013niqe}, metrics chosen in advance, give $\rho=-0.02$ and $-0.12$, three exploratory deep-network scores~\cite{ke2021musiq,chen2024topiq,wang2023clipiqa} reach at most $\rho=0.25$, the wrong sign (no CI lies below 0, and CLIP-IQA's lies above it), and \ours's own confidence ranks it weakly, if at all ($\rho=-0.15$, CI [$-0.38$, $0.11$]; Sec.~S8).
Nor do the frozen generic scores flag the scenes on which the non-expert and the reference differ by over 10\% (post hoc; every AUROC interval spans 0.5; Sec.~S8).
Both tests rest on human counts that repeat only loosely across re-shoots (\cref{tab:repeat}), which can hide a weak signal; on the legacy crops, four generic scores flag defocus and motion blur within a fabric about as well as \ours's confidence, but none beats chance on generated texture (exploratory; Tab.~S3).
The frozen quality gate, which sends scenes of low confidence or resolvability to GPT-6.1-sol, barely changes \ours's error (\cref{tab:main150}; Secs.~S8 and S10).

\subsection{Capture stress and the ruler}
\label{sec:exp_quality}

\begin{table}[t]
\centering
\footnotesize
\setlength{\tabcolsep}{5pt}
\begin{tabular}{@{}lcc@{}}
\toprule
Degradation, levels & \multicolumn{2}{c}{Stable scenes (\%)} \\
(1\,u $=$ 0.001\,inch) & \ours & FFT peak \\
\midrule
\csname @@input\endcsname tables/quality76_main_rowsB.tex
\bottomrule
\end{tabular}
\caption{Capture stress without labels: share of the 76 scenes whose count stays within 5\% of the clean count (FFT peak: no estimate counts as unstable); bold: the more stable counter at each level.}
\label{tab:quality}
\end{table}

\begin{figure*}[t]
  \centering
  \includegraphics[width=\textwidth]{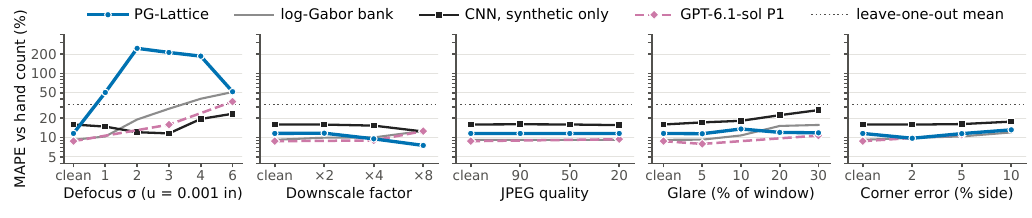}
  \caption{Defocus makes \ours over-count; the other methods change less. MAPE against the hand counts of the 13 curated legacy crops\LEG (log scale; lower than on \dataset, Tab.~S1) under five degradations, with byte-identical pixels for every method and one VLM run per condition; dotted: the leave-one-out mean.}
  \label{fig:vlmdeg}
\end{figure*}

\paragraph{Motion blur, corner error, defocus and too few pixels break counts.}
We apply eight physically scaled degradations to every scene and check, without labels, whether each count stays within 5\% of the clean one (\cref{tab:quality}): for \ours, motion blur of 5\,u and a corner error of 5\% of the side each destabilize about half of the scenes, while JPEG and noise at their milder levels and shadow at both levels keep nearly all stable.
On the legacy crops, where every method sees byte-identical degraded pixels and is scored against the hand counts (\cref{fig:vlmdeg}), defocus of 2--4\,u makes \ours lock onto residual high-frequency artifacts and over-count severalfold (the plain FFT peak alike), while log-Gabor, the CNN and GPT-6.1-sol stay far lower.
In simulated re-captures of \dataset and legacy crops, counts collapse below about 3\,px per yarn, and through a phone pipeline extra jumps of more than 10\% persist up to about 7 (Sec.~S12), so a capture app should aim for 7\,px per yarn, refuse fewer than 4 and never up-sample (\cref{sec:exp_genai}).

\paragraph{On these photos, the glass's ruler can stand in for corner annotation, at a small cost.}
The ruler (\cref{sec:ruler}) calibrates all but one scene, a median 0.19\,mm from human corners, and keeps 92\% of scenes within 5\% of the annotated-corner count, against 42\% for a classical segmentation (Sec.~S9); against the reference its crop raises \ours's error on those scenes from 35.3 to 37.9\% ($+2.6$ points; paired, post hoc, CI [0.0, 6.7]). %

\subsection{Enhancement does not reliably help the count}
\label{sec:exp_genai}
In a pre-registered study (Sec.~S13) we down-sampled ($\times$4, $\times$8, $\times$4+JPEG) the legacy crops and 53 large, confidently counted \dataset crops (40 scenes; copies included, which narrows the CIs), and enhanced them with bicubic interpolation, a regression super-resolution network or one of five generative restorers (\eg~\cite{wang2021realesrgan,wang2026seedvr2,yue2025invsr}).
As pre-registered before any human count, the 53 crops are scored against each method's own clean-crop count (secondary reference; primary: Sec.~S13).
The planned baseline, bicubic up-sampling, failed, as \ours locks onto interpolation replicas of the yarn frequency; so, in a change decided before any generative output was scored, we compare each enhancer with simply counting the degraded crop at its own resolution.
Enhancement did not help on balance: of 27 generative comparisons (paired 95\% CIs; Sec.~S13), 3 improved \ours and 3 worsened it, the CNN got worse in 22, two generative restorers added off-lattice periodicity, and every generative output scored a mean MUSIQ~\cite{ke2021musiq} at or above the real high-resolution photo's: perceptual quality does not certify measurement fidelity~\cite{blau2018perception}.
A \emph{native-resolution guard}, designed after this failure and set on synthetic crops, recounts at the native side when the claimed count exceeds the content bandwidth: it removes the bicubic failure (legacy $\times$8: 301.7 to 6.0\%\LEG) and never fires on a clean crop, but misses generative texture and turns most 6\,u defocus counts into under-counts (Sec.~S14).

\section{Limitations and Societal Impact}
\label{sec:impact}

\paragraph{Limitations.}
No count was checked on the cloth, so which octave is right, and how close any count comes to the true density, remain open; a recount on the cloth and a counting convention that annotators share are the next step.
The photos come from one mill and one glass, and a third of the scenes exist only as a quarter-width copy.
One textile-trained annotator, with no second expert to check them, is the reference and one non-expert the human baseline, both from one recruitment round; scoring against the reference rather than the pre-registered mean of both is a declared deviation, and some rankings depend on it (Tabs.~S6 and S7, Sec.~S7); and the ``check carefully'' note may have primed both toward the half-or-double question (\cref{sec:data}).
Comparisons with hand counts under degradation and enhancement rest on 13 legacy crops with noisy labels.
VLM results are a snapshot at temperature 0 and the lowest reasoning setting; higher settings may do better.
Color-woven checks add spectral peaks; satins other than 5-end ones, basket weaves, jacquards and knits fall outside the lattice model; the ruler fails below about 120\,px per inch; and the phone's optical zoom without the glass is untested.

\paragraph{Societal impact.}
Small units in India's textile sector~\cite{ibef2026textiles} check density with a pick glass costing a few hundred rupees, against Rs~8{,}000--35{,}000 for a thread counter~\cite{indiamart2026threadcounter}.
A counter that met the tolerance could screen more places per piece; none does yet, and we have not measured time saved or adoption.
\ours runs offline on a laptop CPU (3.5\,s per crop) and our tuned Molmo2-4B has open weights a mill could host on one GPU, while GPT-6.1-sol needs a paid API call per photo, sends factory images to a third party whose model can change silently, and needs mobile internet, which women in South Asia are far less likely than men to use~\cite{gsma2025gendergap}.
Monitoring individual weavers is an excluded misuse, and a model card will report error per fabric type and quality bin~\cite{mitchell2019modelcards}.
The manual count of IS~1963 and ASTM~D3775~\cite{bis1981is1963,astm2017d3775} remains the arbiter: a deployed system should show the yarns it counted, refuse up-sampled or ``enhanced'' input and abstain when unsure, and because a halved count makes a compliant piece look 50\% below specification, a count near half or double must go to a manual recount before it can affect a piece's acceptance or a weaver's pay.

\section{Conclusion}
\label{sec:conclusion}
Through the inspector's own pick glass, thread counting becomes a metric-image problem: the glass's scales give the window, and a $\pm$5\% tolerance says when a photo is good enough.
On \dataset no method agrees with a textile-trained reference within it; a small open VLM that we tuned only on synthetic images is off by about a quarter, level with the best frontier model and below a non-expert annotator; our training-free counter most often repeats a count within 5\% but can lock onto half or double the yarns; even the reference repeats only to a median 9\% across re-shoots, so the image quality that matters is whether a count repeats; and generic quality scores chosen before labeling do not rank the counter's error.
In practice, capture about 7\,px per yarn (never fewer than 4), never up-sample or enhance, compare several photos, and recount on the cloth any count they disagree on or that lies near half or double the specification.

{
    \small
    \bibliographystyle{ieeenat_fullname}
    \bibliography{paper_refs}
}

\end{document}

\maketitle

\noindent This supplement collects the detail that does not fit in the 8-page paper. Section, table and figure numbers without an ``S'' refer to the main paper. Numbers marked \LEG are from the 13 hand-counted legacy windows (26 counts), scored under the one crop rule of Sec.~3. Code, predictions and every script named below are at \codeurl{} (\texttt{tools/} and \texttt{tables/} under \texttt{paper/}); the photos, the 13 legacy crops, the human counts and the score files computed from them (\texttt{tables/main150*.csv}, \texttt{strata150\_mape.csv}) will follow after review. An extended supplement, to be added to that repository, will hold verbatim what this supplement no longer prints, among it every legacy method and prompt with the paired tests, every cascade policy, every CNN variant with the pre-registered augmentation ablation, and further studies (VLM measurement scaffolds in full, the synthetic stress benchmark, window segmentation, the label-exact augmentation checks).

\section{Task definition and benchmark protocol}
\label{sec:supp_task}
\paragraph{Measurand and metrics.}
$h$ ($v$) counts the individual threads actually present that cross a horizontal (vertical) 1-inch segment inside the window (ISO~7211-2): a plied yarn counts once, and an edge yarn counts if its center is inside. The annotator guide asked for one dot per individual yarn, with worked cases for two-color (a light and a dark float side by side are two yarns), twill and check fabrics, but did not address plies or same-colored neighboring yarns that work as a pair (Sec.~5.2).
Besides the metrics of Sec.~5, the scorer, fixed before any human count, reports the median of $e=(\hat y-y)/y$ (bias) and $1.4826\,\mathrm{MAD}(e)$ (spread); against the mean of the two annotators the bias is $-2.4$\% for \ours, $-21.3$\% for GPT-6.1-sol P1, $-78.8$ to 0.0\% for the seven VLMs' P1 answers and $+23.0$\% for the synthetic CNN (spreads: \texttt{tables/main150.csv}).

\paragraph{Splits and leakage.}
The legacy crops and \dataset are test-only, except for methods scored out of fold on \dataset: the CNNs fine-tuned on human counts and the learned octave selector (\cref{sec:supp_octave}), trained on the non-expert's counts unless stated, and the pseudo-label CNN, trained on \ours's counts of 28 scenes and scored on the other 48 (\cref{sec:supp_cnnaug}).
Otherwise counters and gates are fitted on synthetic data only, with six disclosed exceptions: the octave check's threshold also used the 13 legacy crops and label-free agreement between copies of the 150 photos (\cref{sec:supp_octave}); the ruler's detector was developed on the 150 photos without counts (\cref{sec:supp_ruler}); the resolvability check and the native-resolution guard were designed after the defocus failure on the legacy crops and the bicubic failure on the legacy and R-HR crops (\cref{sec:supp_cascade,sec:supp_genai,sec:supp_guard}); the CountGD variant was chosen on the legacy crops; and the combined counter's rule family was motivated by the legacy degradation results.
Fine-tuning folds keep all files of a scene together. Folds grouped only by scene let re-shoots of one patch straddle a split, so the CNN of Tab.~1 uses folds grouped by the 44 patches (post hoc on the non-expert's counts; rule~7 of the plan below pre-specified a patch-grouped version of the scene-grouped run, same settings, as a sensitivity row; that row was not run, and the patch-grouped runs reported here were trained on the non-expert's counts, not on the mean as that run was, a deviation); its ImageNet initialization was chosen on the non-expert's counts (against the mean of both, computed afterwards: \FNcnnFabImagenetMape\% \vs \FNcnnFabSynthMape\% with synthetic initialization). The scene-grouped run of \cref{tab:supp_all150,tab:supp_prereg} (\FNcnnSceneMape\% against the mean) is optimistic twice over: re-shoots can straddle its folds, and it was trained on the mean, the reference it is scored against (30.6 and 22.5\% against either annotator alone).

\paragraph{Augmented copies.}
The augmented set holds 843 usable copies of the 150 photos (853 on disk; \texttt{aug1016/}), made by an earlier preparation notebook with 90$^\circ$ rotations, perspective, shift and scale, brightness and contrast, noise and JPEG; each copy inherits its source's counts and scene, and folds follow the source.
It is never scored against human counts: a copy repeats its scene's human count and shares that count's errors, so scoring copies would add rows and narrow intervals without adding evidence (the effective sample stays 76 scenes of 44 patches), and more test evidence needs new scenes and patches, ideally counted on the cloth.
It serves for training and for one label-free check, the kind of test Tab.~3 runs on controlled capture changes: on the copies \ours keeps 92.9\% of its counts within 5\% of its count on the original, against 95.4\% when the original is only re-rectified and rotated (exploratory, \texttt{aug1016/invariance}).
Training on it did not help: the ensemble variant ens-v1b, whose CNN also trains on it, is not better than the ensemble (chosen post hoc, then scored blind; record and scores: ``Pre-registration'', below), and in the six scene-grouped CNNs trained with it or with label-exact augmentation (\cref{sec:supp_cnnaug}), not blind tests as these runs trained on the mean, the MAPE against the mean is 15.5--16.8\%, \vs 15.6 and 15.8\% without (\texttt{tables/main150.csv}, rows \texttt{cnn-cv-*}).

\paragraph{The ensemble.}
The CNN--VLM ensemble, a per-axis geometric mean of the patch-grouped CNN and GPT-6.1-sol P1 (\texttt{PREREG\_ens\_v1.json}), is post hoc; its record was frozen before the reference annotator's counts were opened (a pre-registered record, below), and it is scored blind; snapping \ours to the CNN's octave was tried and was worse (\FNsnapCnnMape\% against the mean). The ensemble's CNN alone, a secondary endpoint of that record, scores 28.1\% against the reference annotator, $+1.0$ points [$-4.1$, $+6.3$] \vs GPT-6.1-sol P1.
Against the mean of both annotators the ensemble improves on GPT-6.1-sol alone by \FNctlEnsMinusGpt{} points \FNctlEnsMinusGptCI, but so does combining GPT-6.1-sol with the synthetic-only CNN, which never saw a human count (\FNctlCnnZeroMape\% \vs \FNctlEnsMape\% for the ensemble; \FNctlCnnZeroMinusGpt{} \FNctlCnnZeroMinusGptCI; with Qwen3.8-Max-Prime, Gemini 3.8 Flash or \ours, $-0.2$ to $-1.4$, every CI spanning 0; post hoc). A geometric mean hedges between the annotators' octaves, so the ensemble's gain against the mean is not evidence that it learned from human counts.

\paragraph{Pre-registration.}
The plan for the reference annotator's counts and its addendum~1 were fixed after the non-expert's counts had been seen but before the reference annotator's were opened (available on request; SHA-256 prefixes \texttt{9ae1d9caab117b97} and \texttt{586b606c38d6f0f8}), so in the terms of main Sec.~5 their rules govern blind tests.
Rule~2 keeps the mean of both counts as the primary reference and adds a sensitivity reference without every axis on which the annotators differ by $|\log_2(\text{ratio})|>0.5$ (\cref{tab:supp_disputed}); rule~3 scores against each annotator alone; rule~4 fixes in advance how Sec.~5.2 is worded for each outcome; rule~5 adds the log-ratio endpoint; rule~6 adds patch-clustered CIs of every headline number (\texttt{tables/main150\_fabric.csv}; \eg \ours [30.1, 40.3] against the mean, scene CI [30.7, 39.6]); addendum~1 fixes the re-shoot analysis (better-of-two $h$/$v$ assignment) and its interpretation (\cref{sec:supp_iaa}).
Records frozen after the non-expert's counts but before the reference annotator's (this plan, the ensemble, the two below and the octave rules of \cref{sec:supp_octave}) are called pre-registered records here, and the methods they fix are scored blind against the reference. Of the two further such records, a variant of the ensemble whose CNN also trains on the augmented copies (above; chosen post hoc on the non-expert's counts, on which it had looked better, 23.8 \vs 25.5\%, $-1.7$ [$-4.2$, $+0.3$]; \texttt{PREREG\_ens\_v1b.json}; a secondary blind test, ens-v1 staying the primary record) reaches 26.4\% against the reference annotator (both counts within 5\% on 1 of 76 scenes), $-0.7$ points [$-3.8$, $+2.4$] \vs GPT-6.1-sol P1 and $+1.0$ [$-1.2$, $+2.8$] \vs the ensemble, so it is not better (its record placed it in the main text only if it beat the ensemble; main Sec.~4.4 reports the null anyway), and the second, a recipe for a further ensemble cross-validated on the final labels (ens-v2), was not run.
\Cref{tab:supp_prereg} reports rules 2, 3 and 5 for the 20 rows above its last block, scored as the plan specifies, and none changes a conclusion of the main paper: without the disputed lines every error against the mean changes by $-6.2$ to $+0.1$ points with the order of the methods nearly unchanged (Spearman $\rho=0.97$), \ours (29.9\%) still scores worse than the mean-density guess (25.6\%), and the annotators still differ by a median 7.0\% on the remaining lines; against either annotator alone no method has both counts within 5\% on more than 8 scenes (octsel-v1, \cref{sec:supp_octave}: 9 against A2); and 14 of the 20 methods are closer than the non-expert (37.3\%) to the reference. Its last block, added post hoc, scores the untuned counting specialists, Molmo2-4B and GPT-6.1-sol with the insets of \cref{sec:supp_prompts} the same way.
Against the non-expert (A1), GPT-6.1-sol P1 minus the reference annotator's counts is $-12.1$ points [$-23.4$, $-2.6$] in MAPE and $-0.06$ [$-0.17$, $+0.04$] in mean $|\log_2|$ (addendum~1, rule~4).
Rule~5's secondary endpoint, the octave check minus \ours in mean $|\log_2|$ against the reference annotator, is $-0.021$ [$-0.043$, $-0.003$].
Rule~1's histogram of $\log_2(\text{A2}/\text{A1})$ over the 152 lines, in half-octave bins centered on $-1.5,\dots,+1.5$, holds 1, 29, 12, 62, 24, 16 and 8 lines; 44 of the 72 disputed lines (33 scenes) are among the 64 on which \ours reads half or double the non-expert.

\paragraph{Freezing.}
Before any human count the scorer was extended to score per scene and to offer a patch-clustered bootstrap, with its pre-registered metrics (both within 5\%, both within 10\%, median signed error, robust spread) unchanged (\texttt{-{}-unit image} reproduces the earlier outputs), and the code and predictions of the pre-registered methods, among them \ours, and the cascade's decisions were frozen (md5 prefixes: scorer \texttt{5879ef44}, counter \texttt{3d0520d6}, decisions \texttt{57d49a4d}; every checksum: \texttt{CHECKSUMS.md} in the repository). Methods trained on human counts ran out of fold afterwards; the CNN--VLM ensemble and both octave rules were frozen before the reference annotator's counts were opened; VLM prompts contain no test examples and are logged verbatim with model, date and repeats.

\section{Legacy crops}
\label{sec:supp_legacy}\label{sec:supp_crop}
The 13 crops show 10 plain and three 2/1 twill fabrics; their labels are noisy: one is a range, three twill warp counts are derived by formula ($a{+}2b$, while on a6 and r8 \ours, the log-Gabor bank and the CNN return $a{+}b$), four images are marked unclear, and none records the orientation.
Every method is scored on the same pixels (\texttt{vlm/crops/legacy}): the legacy crop with the dark frame trimmed by a per-side dark-band detector and resized to a square; slivers of about 3\% of frame remain on four crops (they bias spectral counts on those axes by about $+2$--3\%), and for every modern method these crops are within 1 MAPE point of the best of the earlier crop rules (\texttt{legacy13\_crop\_sensitivity.csv}).
\Cref{tab:legacy_unified} gives the main rows; no paired difference between \ours and GPT-6.1-sol P1 is significant at $n=13$ ($+2.8$ points [$-5.1$, $+13.2$]).
VLM answers are not deterministic at temperature 0: over three repeats on all 20 legacy crops (13 of them hand-counted), the mean per-image spread is 4.9\,\TPI for GPT-6.1-sol and 14.9 for Gemini 3.8 Flash (\texttt{vlm/RESULTS.md}).

\begin{table}[t]
\centering
\scriptsize
\setlength{\tabcolsep}{2pt}
\resizebox{\linewidth}{!}{%
\begin{tabular}{@{}lcccrr@{}}
\toprule
Method (P1 unless noted) & MAPE [95\% CI] & $\le$5\% & $\le$10\% & s/img & \$/img \\
\midrule
Leave-one-out mean & 32.5 [25.6, 39.6] & 3/26 & 5/26 & -- & 0 \\
\midrule
\ours (CPU) & 11.6 [3.1, 23.1] & 19/26 & 21/26 & 3.5 & 0 \\
Log-Gabor bank (CPU) & 9.2 [3.9, 15.2] & 18/26 & 21/26 & 4.0 & 0 \\
FFT max near-axis peak (CPU) & 18.2 [6.0, 32.8] & 17/26 & 20/26 & 1.8 & 0 \\
CNN, synthetic only & 16.0 [10.8, 21.8] & 4/26 & 10/26 & 4.4 & 0 \\
CNN, + pseudo-labels (B0, V4) & 10.7 [7.0, 14.9] & 11/26 & 16/26 & -- & 0 \\
CountGD (pre-declared primary) & 341.8 [254.7, 432.8] & 0/26 & 0/26 & 2.9 & 0 \\
CountGD, cells $\to$ lattice$^\ddagger$ & 18.3 [10.7, 27.0] & 9/26 & 12/26 & 2.6 & 0 \\
Molmo2-8B / P4q & 79.9 / 24.9 & 0 / 6 & 0 / 8 & 0.99 / 10 & 0 \\
Molmo2-4B & 82.9 [80.4, 85.4] & 0/26 & 0/26 & 0.27 & 0 \\
Molmo2-4B + synthetic LoRA & 18.4 [10.6, 29.1] & 7/26 & 9/26 & 0.19 & 0 \\
\midrule
GPT-6.1-sol & 8.8 [4.5, 13.8] & 16/26 & 21/26 & 16 & 0.011 \\
GPT-6.1-sol, high reasoning & 14.1 [7.1, 22.3] & 17/26 & 19/26 & 97 & 0.047 \\
Gemini 3.8 Flash & 19.1 [12.7, 25.6] & 5/26 & 6/26 & 11 & 0.005 \\
Qwen3.8-Max-Prime & 20.5 [15.3, 25.5] & 1/22 & 6/22 & 31 & 0.023 \\
Gemini 3.1 Pro & 40.8 [27.2, 57.0] & 5/26 & 8/26 & 9.0 & 0.014 \\
Claude Fable 5.1 & 34.9 [29.6, 40.3] & 0/26 & 1/26 & 9.6 & 0.041 \\
Qwen3.8-27B (open) & 61.1 [49.2, 72.2] & 0/26 & 0/26 & 2.6 & 0.0004 \\
Gemma-4-26B-A4B (open) & 54.2 [47.7, 61.6] & 0/26 & 0/26 & 4.4 & $<$0.0001 \\
\bottomrule
\end{tabular}}
\caption{Legacy test\LEG: 13 crops, 26 counts, axis-matched, all methods on the same canonical crops (\cref{sec:supp_legacy}). MAPE with an image-bootstrap 95\% CI; $\le$5\%, $\le$10\%: counts within 5\%, 10\% (parse failures not imputed). s/img: CPU rows median on an M2 Max laptop, VLM rows mean API wall-clock, Molmo2 and CountGD on an RTX 4090. $^\ddagger$Post hoc. Every prompt (P1--P4q), MAE, plain-weave MAE and the paired tests: \texttt{tables/legacy13\_unified.csv} and \texttt{legacy13\_paired.csv} in the release and the extended supplement.}
\label{tab:legacy_unified}
\end{table}

\section{Annotation protocol and agreement}
\label{sec:supp_iaa}
Two external annotators, each hired after a public screening test on two legacy crops (r2, a5; \cref{tab:supp_iaa}, top), counted all 76 representatives from scratch and independently in the tool of Sec.~3 (v2.2.2): A2, with textile training, is the reference; A1, without, is the non-expert.
The counting view showed no machine dot, count or weave guess; on 38 of the 56 scenes outside an earlier 20-scene subset (\cref{sec:supp_scenes}) it showed a ``check carefully'' (26) or ``take care'' (12) badge with a reason (\eg that the spectral estimators disagree on $v$ by 100\%) but no count, and the four worked examples of the guide showed \ours's dots on scenes 27, 65, 93 and 110 (scores without them: \cref{tab:supp_iaa}).
A photo could be marked done only when its weave and a confidence were set and each guide line had at least 10 dots spanning at least 75\% of the line with no gap above three dot spacings, waived for a line marked partially occluded or explained in a note; no line of either annotator fell under the 75\% span (0 partially counted lines of 152 per annotator), so no line is left out of any score.
\Cref{tab:supp_iaa} gives the agreement on all 76 scenes: MAE, Bland--Altman bias with 95\% limits~\cite{bland1986agreement}, ICC(2,1) for absolute agreement~\cite{shrout1979icc}, median difference and weave $\kappa$~\cite{cohen1960kappa}.

\paragraph{Re-shoots (post hoc).}
The 41 same-patch pairs come from 24 patches, so intervals resample patches. With the better-of-two $h$/$v$ assignment fixed in addendum~1 (\cref{sec:supp_task}; the measured rotation of the photos gives the same median and \FNrsPairsSiftFive{} instead of 8 non-expert pairs within 5\%), the number of pairs within 3, 5 and 10\% on both axes is 4, 8 and 10 for the non-expert (A1), 7, 12 and 21 for the reference annotator (A2), 4, 9 and 17 for their mean, 24, 24 and 26 for \ours, 9, 11 and 17 for GPT-6.1-sol P1 and 10, 13 and 21 for the CNN--VLM ensemble (medians: Tab.~2; the non-expert's \FNrsWorseMedian\% has the patch-resampled CI \FNrsWorseMedianCI; by resolution, post hoc and confounded with the fabric as the two groups are different patches, A2's median is \BNrsNGfull\% between full-resolution and \BNrsNGquarter\% between quarter-resolution photos, A1's \BNrsSAfull\% and \BNrsSAquarter\%).
As the two annotators' same-photo median difference, \BNsamePhoto\%, exceeds 10\%, addendum~1's third case applies: the per-photo human reference is too noisy to test $\pm$5\%. As the rule requires, main Sec.~3 says so, the abstract states that one photo cannot yet be shown to meet $\pm$5\% (citing the reference's re-shoot repeatability), the main paper reports agreement with the reference rather than accuracy, and every method is also scored against each annotator alone (\cref{tab:supp_prereg}).

\paragraph{Badge split (post hoc).}
Against the mean of the two annotators, \ours reads half or double on \FNbadgeOct{} of the 38 scenes that showed a badge and on \FNnoBadgeOct{} of the 38 others (Fisher $p$ = \FNbadgeOctP), and the non-expert's photo counts lie on average \FNrsBadgeDist\% from their patch median on badge scenes against \FNrsNoBadgeDist\% on the others (Mann--Whitney $p$ = \FNrsBadgeDistP), so the badge did not make that annotator less consistent. %

\begin{table}[t]
\centering
\footnotesize
\setlength{\tabcolsep}{3pt}
\resizebox{\ifdim\width>\linewidth\linewidth\else\width\fi}{!}{%
\begin{tabular}{@{}lcc@{}}
\toprule
Two annotators (A1, A2), from scratch & $h$ (\TPI) & $v$ (\TPI) \\
\midrule
Screening r2: legacy hand count & 99 & 69 \\
\quad answers of A1 / A2 & 97 / 99 & 69 / 68 \\
Screening a5: legacy hand count & 47 & 49 \\
\quad answers of A1 / A2 & 46 / 48 & 50 / 49 \\
\midrule
Scenes counted by both & \multicolumn{2}{c}{76} \\
MAE between annotators & 27.75 & 25.99 \\
Bland--Altman bias [95\% LoA] & $-0.69$ [$-77.7$, $+76.3$] & $-0.07$ [$-69.7$, $+69.6$] \\
ICC(2,1), absolute agreement & 0.201 & 0.284 \\
Median $|$difference$|$ (\%), all & 32.9 & 30.1 \\
\quad badge / no badge & 47.9 / 28.3 & 35.3 / 22.4 \\
Scenes with both axes within 5\% & \multicolumn{2}{c}{\BTannBothFive{} of 76} \\
Weave $\kappa$ & \multicolumn{2}{c}{$-0.01$} \\
Warp direction & \multicolumn{2}{c}{not determinable, not scored} \\
\midrule
\ours MAPE, all / without 27, 65, 93, 110 & \multicolumn{2}{c}{35.1 / 35.3} \\
Median time per photo, A1 / A2 (min) & \multicolumn{2}{c}{3.8 / 9.6} \\
\bottomrule
\end{tabular}}

\caption{Label checks and agreement of the two annotators, A1 and A2. Top: raw screening answers against the legacy hand counts (axis-matched). Second block: agreement on the scenes counted by both, also on the scenes with and without a count-free badge; median $|$difference$|$ per axis, relative to the mean of the two (pooled over both axes: \BNsamePhoto\%, as in the main paper, and \BTnoBadgeMed\% on the \BTnoBadgeN{} scenes without a badge); warp direction: A1 blank on 75 scenes and ``unknown'' on one, A2 ``unknown'' on all 76. Bottom: sensitivity to the four scenes whose machine dots appeared in the guide (rounded from a two-decimal value, so it can differ from \cref{tab:supp_meanref} in the last digit; without them, as pre-specified, the 72 scenes give ICC(2,1) 0.221 / 0.285, median $|$difference$|$ 32.9 / 33.1\% and weave $\kappa$ $-0.02$), and annotation time.}
\label{tab:supp_iaa}
\end{table}

\paragraph{Paired tests against the reference (post hoc).}
The CNN--VLM ensemble is also closer to the reference than the non-expert (MAPE \BNensMinusHumNG{} points, CI \BNensMinusHumNGci; $|\log_2|$ \BNlogEnsMinusHum, CI \BNlogEnsMinusHumci), while on MAPE \ours cannot be separated from the non-expert (\BNoursMinusHumNG, CI \BNoursMinusHumNGci).
Patch-clustered 95\% CIs (resampling the 44 patches): GPT-6.1-sol P1 minus A1, [$-19.6$, $-1.5$] MAPE points, and Molmo2-4B + LoRA minus A1, [$-21.0$, $-3.6$]; minus the leave-one-out mean, GPT-6.1-sol [$-13.0$, $+1.0$] and Molmo2-4B [$-14.5$, $-1.4$] (on $|\log_2|$, [$-0.08$, $+0.14$] and [$-0.19$, $-0.01$]); Molmo2-4B minus GPT-6.1-sol on $|\log_2|$, [$-0.25$, $-0.02$].

\section{\ours: scoring function and more examples}
\label{sec:supp_lattice}
\paragraph{Scoring function} (details of Sec.~4.2).
Pre-processing uses local contrast normalization $(I-G_{S/40}*I)/\sqrt{G_{S/16}*(\cdot)^2}$ after projecting color onto the first principal axis of the high-passed RGB vectors and masking specular glare as solid saturated blobs.
$g(\mathbf{p})$ is the whitened score at frequency $\mathbf{p}$ minus the local noise baseline, capped at 15. A hypothesis $H=(\avec,\bvec,W)$ is scored as
\begin{align}
S(H) ={}& g(\avec)+g(\bvec)+\tfrac12[g(\avec{+}\bvec)+g(\avec{-}\bvec)] \nonumber\\
&+\tfrac14[g(2\avec)+g(2\bvec)]+\operatorname*{mean}_{\mathbf{p}\in W} g(B\mathbf{p}) \nonumber\\
&- \tfrac14\!\!\sum_{\mathbf{p}\in P_{\text{sub}}}\!\! g^+(B\mathbf{p}) - \!\!\sum_{\mathbf{q}\in Q_H}\!\! g^+(\mathbf{q}) \nonumber\\
&-E(\avec)-E(\bvec)-A(\theta_a,\theta_b),
\label{eq:score}\tag{S1}
\end{align}
where $B=[\avec\;\bvec]$, $P_{\text{sub}}=\{(\tfrac12,0),(0,\tfrac12),(\tfrac13,0),(0,\tfrac13)\}$ are on-axis sub-harmonic positions, $E(\mathbf{x})=\sum_{n=2}^{8}c_n[g(n\mathbf{x})-g(\mathbf{x})]^+$ with $c_n=\tfrac12$ for $n\le3$ and 1 otherwise, and $Q_H$ is the set of significant peaks ($z\ge7$) finer than $H$ (lattice coordinates $>1.1$) that are neither lattice points nor cosets of $W$.
Because yarns are the finest periodic structure, an unexplained finer peak marks $H$ as a color or weave repeat; the envelope term $E$ penalizes harmonics stronger than the fundamental, and $A=\max(0,|\theta_a{+}\theta_b|-1^\circ)+0.05(|\theta_a|{+}|\theta_b|)$ encodes that a rigid rotation tilts $\avec$ and $\bvec$ by $\pm\theta$.

\paragraph{Fusion and confidence} (details of Sec.~4.2).
The lattice estimate is kept if either of two independent estimators agrees within 6\%: the period of a deskewed projection profile, found with YIN over four bands, and the median FFT estimate of 3$\times$3 overlapping half-size tiles; otherwise the agreeing pair, or the median of the three, is used.
The confidence is a logistic model on no-reference features (yarn-peak strength, hypothesis margin, tile support, estimator agreement, sharpness, glare fraction), fitted on synthetic data only (5-fold AUC 0.84 for ``both axes within 10\%''); the resolvability $\eta$ uses the same sharpness ratio $s$.
The weave label is the selected sub-lattice $W$ if its mean evidence $g\ge2$, else ``unsure''.
Every design choice was made on 300 rendered images with exact labels; the legacy labels were used once, after freezing.

\paragraph{Ablation on synthetic data} (for Sec.~4.2).
On the 300 rendered images with exact labels, each design element added to an intermediate configuration gains within-10\% accuracy: the rotation prior 6 points, the cross terms 3.5, the weave sub-lattice 3.2, the finer-peak penalty 2.5 and fusion 4.8.

\begin{figure}[t]
  \centering
  \includegraphics[width=0.965\linewidth]{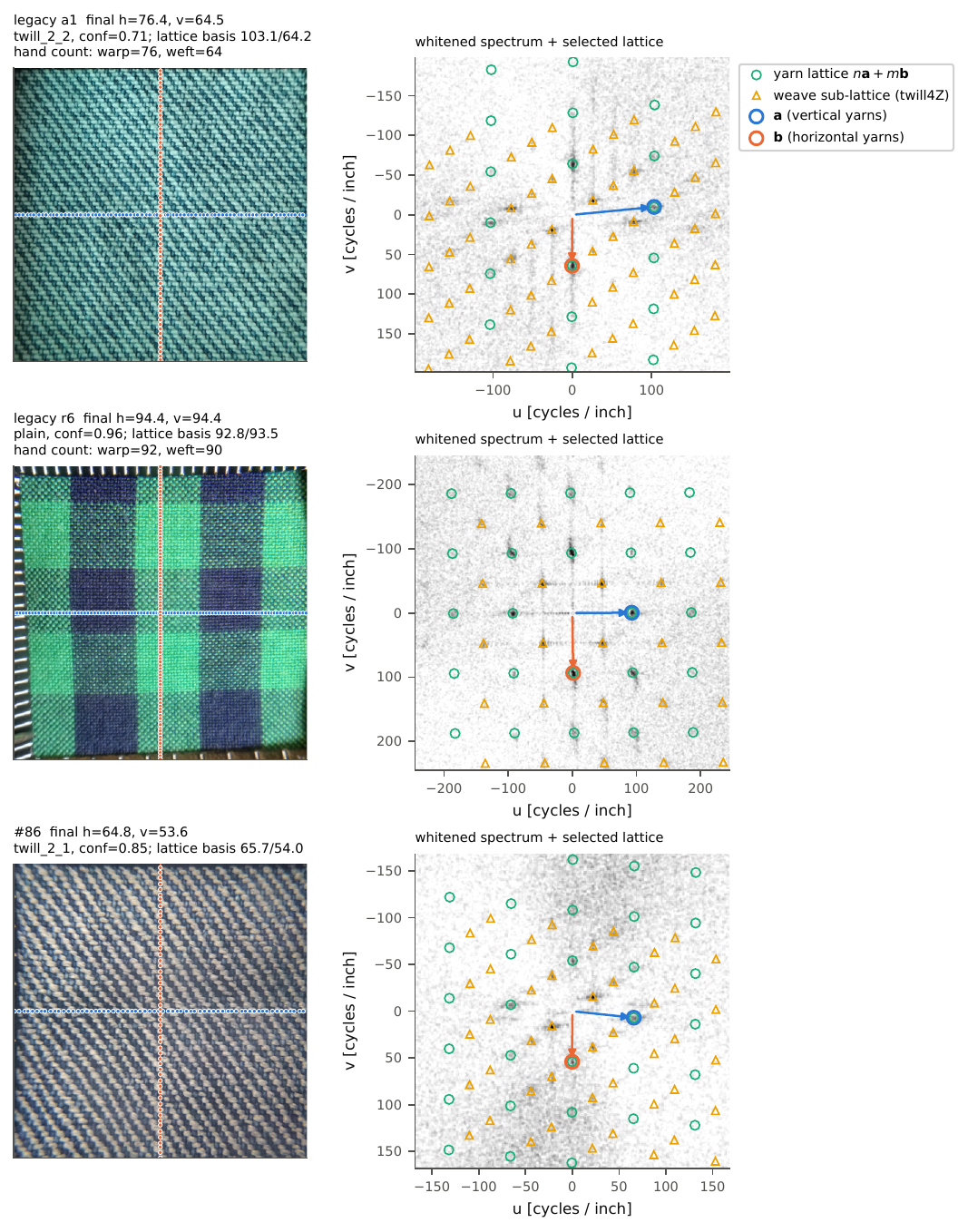}
  \caption{\ours on legacy a1 (2/1 twill; counts within 1\%, warp label formula-derived), legacy r6 (plain-weave check) and \dataset \#86: the crop with the counter's crossings (never shown to the annotators), and the whitened spectrum with the selected yarn lattice $n\avec+m\bvec$ and weave sub-lattice. Twill-diagonal peaks are explained as the weave sub-lattice.}
  \label{fig:supp_lattice}
\end{figure}

\paragraph{Examples.}
\Cref{fig:supp_lattice} shows the counter on further images (legacy panels on the canonical crops).
In the plain-weave check r6 the color repeat creates coarser on-axis peaks; a hypothesis built on them leaves the finer yarn peaks unexplained and is penalized by the $Q_H$ term of \cref{eq:score}.
The failure cases are the blurred low-contrast twill a6, where no weave sub-lattice has enough evidence and the weft locks onto its second harmonic (85 \vs 43; its warp, 31, is the $a{+}b$ reading of the formula label 46), and photo \#55, where the lattice finds 52$\times$42 but the profile and tile estimators agree on the half-frequency float pattern, so fusion returns 26$\times$21 (confidence 0.59); the annotators split (52$\times$41, matching the lattice, \vs 156$\times$123, about three times it), so the human counts do not decide which reading is right.

\section{Synthetic data}
\label{sec:supp_synth}
The renderer draws woven fabric analytically (plain, basket, twills, herringbone, satin; 30--200\,\TPI; yarn-dyed patterns, slubs, skew, density drift), simulates the capture (exposure, vignetting, shadows, LED glare, the dark frame, a random homography, defocus, motion blur, 380--1100\,px/inch, noise, JPEG) and rectifies with jittered corners; its second version also renders the 12{,}000 pointing crops of Molmo2-4B (\cref{sec:supp_lora}).
The CNN (a ResNet-18) is trained on 4{,}800 such crops with AdamW~\cite{loshchilov2019adamw} and resamples each crop to $P$\,px/inch before tiling; its estimate is the median over tiles, three values of $P$ and flips.
The script \texttt{realism\_stats.py} compares capture statistics of 300 synthetic crops with the 150 real crops at 768\,px/inch (median [p90]).
The first renderer version had blown-out glare (p90 clipped-pixel fraction 0.082 \vs 0.001 real), over-saturated color (median HSV saturation 93 \vs 44) and no deep shadows (p90 deep-shadow area 0.00 \vs 0.18).
The second version reduces these to 0.005, 67 and 0.03 (the CNN's 4{,}800 crops are 2{,}400 from each version); real photos still have more and larger deep shadows, and real yarn hairiness is not modeled.

\section{VLM prompts}
\label{sec:supp_prompts}
The seven VLMs of Tab.~1 are GPT-6.1-sol, Claude Fable~5.1, Gemini~3.1~Pro preview, Gemini~3.8~Flash and Qwen3.8-Max-Prime (closed) and Qwen3.8-27B and Gemma-4-26B-A4B (open weights), all queried via OpenRouter (model ids and query dates: code release; prompts: \texttt{vlm/prompts.py}).
\textbf{P1} (verbatim): ``This is a magnified photo of woven fabric seen through a 1-inch pick glass. Count the number of vertical yarns crossing a horizontal line across the full window, and the number of horizontal yarns crossing a vertical line. Answer JSON \{vertical\_yarns:int, horizontal\_yarns:int, weave: plain|twill|satin|other\}''.
\textbf{P2} adds that the image is exactly one inch square and asks the model to identify the weave, count in a small region and scale up, sanity-check, and give the JSON on the last line.
\textbf{P3} states that the image is a patch exactly 1/4 inch wide and tall and otherwise repeats P1; \textbf{P3s} asks P3 on the four diagonal quarter-inch patches and sums the answers.
Every VLM receives the same crop resized to 1536$\times$1536\,px, at temperature 0 and the lowest hidden-reasoning setting each API accepts (on the legacy crops GPT-6.1-sol's high setting was not better: MAPE 14.1 \vs 8.8\%, \cref{tab:legacy_unified}). The 50 full-resolution windows (1838--2032\,px) are thus down-sampled and the 26 quarter-width ones (485--521\,px) up-sampled about threefold (bicubic). Up-sampling does not measurably change the VLMs' error (post hoc): on the 47 full-resolution scenes that also have a quarter-width copy of the same exposure, scored against the reference annotator (A2) with the same counts, the copy minus the full photo changes MAPE by $+0.1$ points [$-4.6$, $+4.7$] for GPT-6.1-sol, $+0.8$ [$-5.0$, $+6.9$] for Claude Fable~5.1, $-1.0$ [$-9.9$, $+8.0$] for Gemini~3.1~Pro, $-2.5$ [$-6.5$, $+1.9$] for Gemini~3.8~Flash, $+2.5$ [$-2.3$, $+7.4$] for Qwen3.8-Max-Prime (41 pairs), $-3.8$ [$-8.3$, $+0.5$] for Qwen3.8-27B and $+3.4$ [$+0.3$, $+7.0$] for Gemma-4-26B (46 pairs), against $+4.1$ [$+0.6$, $+8.6$] for \ours, which counts at native resolution (paired scene bootstrap).
\textbf{P4} (pointing, Molmo2): ``Point to every vertical yarn where it crosses the red horizontal line.'' and the analogous vertical-line prompt, with at most 1{,}536 output tokens.
\textbf{Scaffolds.} A study of four measurement scaffolds on the legacy crops (extended supplement) pre-declared four magnified 0.1-inch insets as primary and tested them against each model's own P1 (Holm over four models): they do not help GPT-6.1-sol ($+2.1$ points [$-2.3$, $+7.5$]) or Gemini 3.8 Flash ($+4.1$ [$-3.1$, $+12.0$]; Holm-adjusted Wilcoxon $p=1.0$ for both; Flash used no hidden reasoning on 51 of 80 scaffold calls, a confound), and lift Qwen3.8-27B and Molmo2-8B ($-39.9$ and $-45.0$ points; $p<0.001$ for both) without beating the leave-one-out mean (Qwen $-11.4$ [$-22.9$, $+0.4$]), Qwen no more than the quarter-inch prompt P3 does ($-0.8$ [$-5.0$, $+2.7$]); on \dataset, against the annotators' mean, they change the MAPE of GPT-6.1-sol, Flash and Qwen by $-2.3$ [$-5.7$, $+1.1$], $+0.4$ [$-3.0$, $+3.7$] and $-39.8$ [$-46.6$, $-32.1$], and against the reference annotator (A2; blind, the insets' counts were exported before any human count) by $-2.2$ [$-6.9$, $+2.6$], $-1.7$ [$-5.3$, $+1.9$] and $-32.4$ [$-40.4$, $-23.1$]: against the reference, GPT-6.1-sol with the insets reaches 24.9\% [19.5, 30.7], level with Molmo2-4B's 25.2\% ($-0.3$ [$-5.1$, $+4.7$]; \cref{tab:supp_prereg}, last block); on the symmetric error it is level with Molmo2-4B ($+0.04$ [$-0.03$, $+0.13$]) and closer than its own P1 answers ($-0.09$ [$-0.15$, $-0.02$]), and it beats the leave-one-out mean on MAPE ($-8.3$ [$-14.0$, $-2.7$]; over patches [$-14.9$, $-1.4$]) but not on the symmetric error ($-0.06$ [$-0.14$, $+0.02$]; over patches [$-0.15$, $+0.04$]), and the non-expert (A1) on both ($-12.4$ [$-19.2$, $-6.0$]; $-0.15$ [$-0.23$, $-0.08$]) (post hoc).

\section{Counting specialists}
\label{sec:supp_spec}
The specialists of \cref{tab:legacy_unified} fail as follows (\texttt{gpu\_results/RESULTS.md}): Molmo2-8B answers a near-constant 12$\times$12 (P1: 79.9\%) or points at evenly spaced templates (P4q: 24.9\%), and CountGD boxes interlacing cells, about three per yarn (341.8\%); its post-hoc cells$\to$lattice readout (18.3\%) is no better than the periodicity estimate that places its exemplars (24.5\%; $-6.2$ [$-17.9$, $+4.2$]).
On \dataset they fail as well: against the reference annotator (A2), Molmo2-8B errs by 81.7\% (P1) and 52.3\% (P4) and CountGD's pre-declared readout by 434.4\% (\cref{tab:supp_prereg}, last block; against the mean: \cref{tab:supp_all150}).
Molmo2-8B runs in bf16 on an RTX 4090; CountGD uses its Swin-B FSC-147 checkpoint.
Protocols of Sec.~4.4: Molmo2-8B answers P1--P3s and points at every yarn crossing a red guide line across the window (P4) or its central quarter inch, counted $\times$4 (P4q); CountGD is prompted with ``yarn'' and three automatic exemplars, and a rule fixed before the run counts its detections in a 1-inch strip two yarn spacings tall (a post-hoc variant reads the lattice spacing from whole-crop detections).

\paragraph{Molmo2-4B with synthetic pointing (LoRA).}
\label{sec:supp_lora}
The generator renders each yarn from a continuous coordinate, so mapping every guide-line pixel back through rectification, sensor resampling and perspective gives the exact crossings. Molmo2-4B is LoRA-tuned (rank 32) on 12{,}000 such crops for 1{,}000 steps (20\% P1, 30\% P4 on the whole inch, 50\% P4 on magnified windows); no real image or label is used, with one run, one seed and no hyper-parameter search.
The checkpoint and the readout were fixed on synthetic validation before any legacy scoring: P1 (7.2\%, the best of four readouts; base model 82.0\%).
On the legacy windows P1 reaches 18.4\% [10.6, 29.1] ($-64.5$ points [$-73.0$, $-52.8$] \vs the base model's 82.9\%), behind \ours (9 \vs 21 counts within 10\%, McNemar $p<0.001$; MAPE $+6.7$ [$-4.9$, $+18.7$]) and GPT-6.1-sol ($+9.6$ [$+2.7$, $+18.5$]); P4s (16.4\%) is post hoc.
The tuned point lists are near-perfectly regular (median gap CV 0.000, against 0.04 for the exact crossings), so the model reads the pitch without locating each yarn.
On \dataset (\cref{tab:supp_all150}), against the mean of the two annotators, the pre-declared P1 readout reaches 28.2\% [24.2, 32.4] (paired $-6.9$ points [$-11.1$, $-2.7$] \vs \ours and $+1.5$ [$-2.5$, $+5.8$] \vs GPT-6.1-sol P1) and the post-hoc P4s readout (four magnified quarter inches, summed) 23.3\% [19.7, 27.4] ($-11.8$ [$-16.0$, $-7.7$] and $-3.4$ [$-7.1$, $+0.5$]).
Its lead over \ours depends on the reference: P1 scores 25.2\% against the reference annotator (A2) alone ($-9.6$ [$-15.7$, $-4.1$] \vs \ours) but 47.6\% against the non-expert (A1) ($+5.9$ [$-1.5$, $+14.3$]).

\section{Capture quality}
\label{sec:supp_quality}
Near-saturated pixels (all channels $\ge$240) exceed 1\% of the window in only 5 of the 76 scenes (maximum 4.8\%), so the 5--30\% glare degradations of Sec.~5.5 extrapolate beyond the captured data.
By the capture strata fixed before labeling (\texttt{tables/strata150.csv}; every stratum: \texttt{tables/strata150\_mape.csv}), \ours's MAPE against the mean of the two annotators is 31.3\% [26.7, 35.4] on the 50 full-resolution and 42.2\% [33.6, 51.0] on the 26 quarter-only scenes, against 35.1\% overall; it is 34.3--39.5\% on the scenes either annotator flagged for glare, blur, shadow, low resolution or perspective, 3.9, 49.0, 29.3 and 30.9\% at $<$4, 4--8, 8--16 and $\ge$16\,px per yarn of the mean count (2, 21, 6 and 47 scenes), 31.3--39.8\% by near-saturated pixels, and 29.2, 34.5 and 44.5\% at window side ratios $\le$1.02, 1.02--1.05 and $>$1.05 (16, 47 and 13 scenes).

\paragraph{Which quality signal predicts counting error?}
We scored five generic no-reference IQA models with pyiqa (BRISQUE~\cite{mittal2012brisque} and NIQE~\cite{mittal2013niqe}, chosen in advance; MUSIQ~\cite{ke2021musiq}, CLIP-IQA~\cite{wang2023clipiqa} and TOPIQ-NR, exploratory) on the image the counter sees, and asked how well each flags a photo whose count is off by more than 5\%, next to the counter's confidence, the resolvability $\eta$ and the frozen gate (\cref{tab:supp_iqa}); so that the confidence is not judged only against its own counter, we use three errors: \ours's, GPT-6.1-sol P1's, and ``no method within 5\%'' (none of the methods compared has both counts within 5\%: four on the legacy crops, \cref{tab:supp_iqa}; three on \dataset, below).
On the non-generative degradations of the 13 hand-counted legacy crops, BRISQUE, NIQE, MUSIQ and TOPIQ-NR reach AUROC 0.57 to 0.76 against 0.74 to 0.75 for the confidence, with no paired difference excluding zero, and within a fabric they flag defocus and motion blur about as well as the confidence (0.79 to 0.95 \vs 0.79 and 0.84; with 13 fabrics, ``about as well'' means indistinguishable at this sample size).
On generated texture no generic score beats chance against the hand counts (AUROC 0.28 to 0.61; every CI includes 0.5 or lies below it), while the confidence keeps 0.73 [0.50, 0.91].
On \dataset, the 76-scene test fixed before any human count (\texttt{tools/iqa150.py}; Sec.~5.4 of the main paper) finds no score that ranks \ours's per-scene error against the mean of the two annotators (Spearman $\rho$: confidence \FNconfRho{} \FNconfRhoCI, BRISQUE \FNiqaRhoBrisque, NIQE \FNiqaRhoNiqe, MUSIQ \FNiqaRhoMusiq, CLIP-IQA \FNiqaRhoClipiqa, TOPIQ-NR \FNiqaRhoTopiq, $\eta$ \FNiqaRhoEta).
Against A2 alone, the main paper's reference (post hoc), the same test gives BRISQUE $-0.02$ [$-0.28$, $0.22$], NIQE $-0.12$ [$-0.35$, $0.12$], MUSIQ $0.07$ [$-0.17$, $0.30$], CLIP-IQA $0.25$ [$0.02$, $0.46$], TOPIQ-NR $0.15$ [$-0.10$, $0.39$], $\eta$ $-0.06$ [$-0.32$, $0.20$] and the confidence $-0.15$ [$-0.38$, $0.11$].
For \ours, which fails on \FNoursFailN{} of 76 scenes, the AUROCs rest on its 2 successes and are uninformative (\eg NIQE 0.78, CLIP-IQA 0.20). In the pre-specified primary analysis (working image, all scenes, three targets, every signal; \texttt{tools/iqa150.py}), $\eta$ flags the scenes on which none of \ours, log-Gabor and GPT-6.1-sol has both counts within 5\% (\FNanyFailN{} failures; AUROC \FNanyFailEtaAuc{} [0.65, 0.93]) and GPT-6.1-sol's failures (\FNgptFailEtaAuc{} [0.67, 1.00]), the exploratory TOPIQ-NR reaches 0.73 [0.60, 0.85] on the latter, BRISQUE, NIQE and \ours's confidence stay near chance (0.46--0.64), and MUSIQ, CLIP-IQA and the gate reach 0.45--0.69 (every CI spans 0.5); this rests on \FNgptBothFive{} GPT-6.1-sol successes against \FNgptFailN{} failures (7 against 69 for the first target), resolution alone flags GPT-6.1-sol's failures at only \FNgptFailQuarterAuc{} (quarter resolution) and \FNgptFailPpiAuc{} (px per inch), and within the 50 full-resolution scenes (pre-specified secondary) $\eta$'s \FNgptFailEtaAucFull{} rests on \FNgptSuccessFull{} successes; against A2 alone (post hoc; 61 and 68 failures) the values fall to $\eta$ 0.54 [0.35, 0.72] and 0.59 [0.35, 0.82] and TOPIQ-NR 0.51 [0.25, 0.75], while \ours's confidence rises to 0.68 [0.49, 0.85] and 0.70 [0.46, 0.89] (every CI spans 0.5).
Post hoc, for flagging the 64 scenes on which the annotators differ by more than 10\% on an axis (12 do not), every AUROC interval spans 0.5 on the working image, the primary input of the frozen test; on its pre-specified secondary 1024-px input, MUSIQ and TOPIQ-NR reach 0.73 [0.56, 0.86] and 0.72 [0.52, 0.91] (lower quality, more disagreement; Mann--Whitney $p=0.014$ and 0.015, both 0.14 after Holm over the ten score--input pairs).
In each of eleven post-hoc subsets (full resolution; no annotator flag; annotator confidence $\ge$2 or 3; $\ge$10\,px per yarn; kept by the gate; \ours's confidence in the top half; the best half by BRISQUE, MUSIQ or TOPIQ-NR; all filters together) \ours gets both counts within 5\% on \FNsubsetsOursMaxText{} (\FNsubsetsSizeRange).

\begin{table*}[t]
\centering
\scriptsize
\setlength{\tabcolsep}{3.2pt}
\begin{tabular}{@{}lccccccc@{}}
\toprule
 & \multicolumn{5}{c}{Non-generative degradations of the 13 hand-counted legacy crops} & \multicolumn{2}{c}{Generative restorations} \\
\cmidrule(lr){2-6}\cmidrule(l){7-8}
 & \multicolumn{2}{c}{\ours error, 247 crops} & GPT-6.1-sol P1 & No method & \ours, within & \ours, R-LEG & IQA up \\
\cmidrule(lr){2-3}
Quality signal & $\rho$ & AUROC & AUROC, 117 & within 5\%, 117 & fabric, 247 & AUROC, 191 & vs bicubic \\
\midrule
\ours confidence & $-$0.40\,{\tiny[$-$0.64,\,$-$0.05]} & 0.74\,{\tiny[0.54,\,0.90]} & 0.75\,{\tiny[0.59,\,0.91]} & 0.74\,{\tiny[0.55,\,0.88]} & 0.67\,{\tiny[0.52,\,0.84]} & 0.73\,{\tiny[0.50,\,0.91]} & -- \\
Resolvability $\eta$ & $-$0.53\,{\tiny[$-$0.72,\,$-$0.25]} & 0.71\,{\tiny[0.54,\,0.87]} & 0.67\,{\tiny[0.51,\,0.78]} & 0.64\,{\tiny[0.48,\,0.81]} & 0.80\,{\tiny[0.65,\,0.92]} & -- & -- \\
Cascade gate (frozen) & $-$0.35\,{\tiny[$-$0.55,\,$-$0.13]} & 0.63\,{\tiny[0.54,\,0.72]} & 0.58$^{*}$\,{\tiny[0.49,\,0.68]} & 0.57$^{*}$\,{\tiny[0.49,\,0.70]} & 0.71\,{\tiny[0.59,\,0.83]} & -- & -- \\
\midrule
BRISQUE$^{\dagger}$ & $-$0.36\,{\tiny[$-$0.55,\,$-$0.09]} & 0.63\,{\tiny[0.46,\,0.80]} & 0.65\,{\tiny[0.53,\,0.74]} & 0.63\,{\tiny[0.48,\,0.77]} & 0.70\,{\tiny[0.57,\,0.83]} & 0.48\,{\tiny[0.28,\,0.67]} & 18/27 \\
NIQE$^{\dagger}$ & $-$0.20\,{\tiny[$-$0.40,\,0.04]} & 0.57\,{\tiny[0.42,\,0.71]} & 0.64\,{\tiny[0.55,\,0.72]} & 0.57\,{\tiny[0.49,\,0.67]} & 0.56\,{\tiny[0.40,\,0.71]} & 0.28$^{*}$\,{\tiny[0.13,\,0.44]} & 25/27 \\
MUSIQ & $-$0.37\,{\tiny[$-$0.52,\,$-$0.14]} & 0.66\,{\tiny[0.54,\,0.77]} & 0.69\,{\tiny[0.58,\,0.81]} & 0.69\,{\tiny[0.60,\,0.79]} & 0.70\,{\tiny[0.55,\,0.84]} & 0.45$^{*}$\,{\tiny[0.34,\,0.58]} & 27/27 \\
CLIP-IQA & $-$0.05\,{\tiny[$-$0.29,\,0.28]} & 0.59\,{\tiny[0.42,\,0.71]} & 0.45$^{*}$\,{\tiny[0.34,\,0.56]} & 0.53$^{*}$\,{\tiny[0.38,\,0.62]} & 0.37$^{*}$\,{\tiny[0.24,\,0.52]} & 0.61\,{\tiny[0.47,\,0.75]} & 27/27 \\
TOPIQ-NR & $-$0.48\,{\tiny[$-$0.61,\,$-$0.24]} & 0.76\,{\tiny[0.66,\,0.84]} & 0.67\,{\tiny[0.57,\,0.77]} & 0.76\,{\tiny[0.68,\,0.84]} & 0.68\,{\tiny[0.54,\,0.80]} & 0.53\,{\tiny[0.41,\,0.64]} & 27/27 \\
\bottomrule
\end{tabular}
\caption{\textbf{Which quality signal predicts counting error, and when.} AUROC for flagging a photo with a count off by more than 5\% (higher = better; 0.5 = chance) and Spearman $\rho$ between quality and the mean absolute \% error (negative = useful), with 95\% CIs from a bootstrap that resamples the 13 fabrics. Columns 1--5: blur, down-sampling, JPEG, glare and corner-error versions of the 13 legacy crops, scored against the legacy hand counts (125 of 247, 79 of 117 and 42 of 117 failures). ``No method within 5\%'': none of \ours, log-Gabor, the CNN and GPT-6.1-sol P1 has both counts within 5\%. ``Within fabric'': only conditions of the same crop are compared, so a signal cannot score by recognizing intrinsically hard fabrics. Column 6: outputs of five generative restorers (Real-ESRGAN, SeedVR2, InvSR, FLUX.2 [klein], Nano Banana Pro) on $\times$4, $\times$8 and $\times$4+JPEG legacy inputs, against the hand counts (77 of 191 failures). Column 7: restorer $\times$ condition cells (of 27, legacy and full-resolution \dataset inputs) in which the score of the generative outputs is above bicubic's (paired 95\% CI above zero). Generic IQA (pyiqa) sees the counter's working image (at most 1024\,px, never up-sampled; generated outputs: 768\,px viewing copies). $^{\dagger}$Metric pre-specified; the other generic-IQA metrics and the within-fabric and generative columns are exploratory. $^{*}$Paired 95\% CI of the difference to \ours confidence excludes zero.}
\label{tab:supp_iqa}
\end{table*}

\section{Window and scale from the glass's ruler}
\label{sec:supp_ruler}
\paragraph{Method.}
The frame carries a fine scale (26 ticks, a long one every fifth) on two adjacent edges and a coarse scale (11 ticks) on the other two (Fig.~1(a)).
Ticks are found by a white top-hat on $\max(R,G,B)$, grouped into rows by line voting and indexed on a 1-D projective lattice per row; a Huber least-squares fit of all tick feet gives the homography, one offset per edge and the window side $W$, and a quality score $q$ combines the fit residual, $|W-26|$ and the coverage (step-by-step recipe in the release).
The detector was developed on the 150 photos without labels; the polygons served only for validation.
The coarse/fine pitch ratio $r=2.5$ of the fit and the 26-division prior of $q$ were set from a first look at the same photos; the free fits under \emph{Glass constants} do not use them.

\paragraph{Accuracy.}
A median of 61 of the 74 ticks is found per photo, at both resolutions, and the fitted corners lie a median 0.20 and 0.19\,mm from the human polygon at about 1890 and 485\,px per inch over the 150 files (149 calibrated; \texttt{tables/ruler\_calib.csv}).
The one photo without a calibration has dim ticks on two edges.
The legacy segmentation is a classical pipeline (shadow removal, two-class k-means, a rotated box; re-implemented with a centered fallback), developed on these 150 photos.
On the 76 scene representatives (\cref{sec:supp_scenes}; \texttt{tables/scenes76\_labelfree.json}) the ruler calibrates 75, at a median 0.19\,mm [0.17, 0.22], and keeps 92\% [84, 96] of the scenes within 5\% of the annotated-corner count, against 42\% [32, 53] for the legacy segmentation (McNemar 39 \vs 1 discordant scenes, $p=7\cdot10^{-11}$).
Started from annotated corners moved by 2, 5 or 10\% of the side, it returns its automatic solution for 99, 99 and 97\% of the photos.

\paragraph{Glass constants.}
Fitted freely per photo, the coarse/fine pitch ratio is 2.506 [2.501, 2.510] (median, 149 photos) and the window side 26.00 [25.98, 26.01] fine divisions.
Because the pitch ratio is independent of depth and perspective, the scales are a 1\,mm/2.5\,mm (or 1/25\,inch/0.1\,inch) pair; whether the opening is 26.0\,mm, or the ticks sit above the fabric plane and are imaged smaller, can only be settled by measuring the glass: an opening of 26.0 fine divisions is +2.4\% wide if they are millimeters and +4\% if they are 1/25\,inch, and the loupe's pincushion ($k_1=-0.026$ from the frame curvature of 50 scenes) adds about +2.7\% on center lines.
\ours's spectrum weights the window center and follows about 2.2$\times$ the area-mean factor, so center-line and whole-window counts can differ by 1--3\%.

\paragraph{Robustness.}
After $\times$4 downscaling the ruler keeps all high-resolution photos ($\approx$470\,px per inch) but only 53\% of the low-resolution ones ($\approx$120\,px per inch), the limit stated in Sec.~6; strong sensor noise ($\sigma=20$ gray levels) is its other weak point.

\section{Quality-aware cascade}
\label{sec:supp_cascade}
\paragraph{Calibration and the legacy evaluation.}
The gate's thresholds are fitted on S-cal only, 200 fresh synthetic crops (generator v2, seeds never used before) rendered on the legacy crops' pixel grid and degraded at the nine legacy conditions of Fig.~4 that have GPT-6.1-sol answers, scored by the frozen \ours: they maximize the offline share subject to at most 10\% of kept crops missing 10\% on either axis ($t_c=0.469$, $\eta_{\min}=0.0022$; 70\% of synthetic crops kept at 9.8\% risk).
Resolvability was designed after the defocus over-count of Sec.~5.5 was known; only its threshold is fitted.
All routers were written into \texttt{tools/cascade\_tradeoff.py} before the single legacy evaluation (117 photos $=$ 13 images $\times$ 9 conditions; stored VLM answers on byte-identical pixels): the gate's error is not distinguishable from GPT-only's (13.6\% \vs 13.5\%; $0.0$ [$-4.6$, $+6.3$], Wilcoxon $p=0.24$) at 24\% of its cost; against \ours alone it gains $-24.1$ points [$-42.3$, $-7.3$], concentrated in the defocus rows (per-image Wilcoxon $p=0.17$), and it removes the worst-condition failure of \ours (212\%$\to$32.9\%), but the synthetic risk level does not transfer: 34\% of the legacy photos kept offline miss 10\% on an axis (9.8\% on S-cal).
Of the routers pre-registered there as secondary, escalating where \ours and the log-Gabor bank disagree, taking the median of the three counts or a vote among them beat GPT-only by 2.5 to 3.2 points (uncorrected Wilcoxon $p=0.005$ to 0.013 at $n=13$), while resolvability alone and internal agreement do not (every policy and the threshold sweep: \texttt{tables/cascade\_legacy13.csv}, \texttt{cascade\_sweep.csv}; extended supplement).

\paragraph{Held-out test on the 76 scenes.}
The frozen gate's per-photo decisions (\texttt{tables/\allowbreak cascade\_150\_\allowbreak decisions.csv}) were fixed before any human count, and scoring them was the held-out test of the cascade.
Against the mean of the two annotators (\cref{tab:supp_meanref}), the gate escalates 15 of the 76 scenes and reaches 34.8\% [30.4, 39.5], against \FNoursMape\% \FNoursMapeCI{} for \ours alone: the paired difference is \FNgateMinusOursDiff{} points \FNgateMinusOursDiffCI{} ($-0.7$ [$-2.9$, $+1.4$] against the non-expert (A1) alone and $-0.3$ [$-3.1$, $+2.3$] against the reference (A2)), with the gate better on 9 of the escalated scenes and worse on 6 (Wilcoxon $p=0.60$, the 61 unchanged scenes counted as ties); it stays 8.1 points [4.2, 12.2] behind GPT-6.1-sol P1.

\section{Provenance of \dataset: 150 files, 76 scenes}
\label{sec:supp_scenes}
\paragraph{Audit and file kinds.}
A read-only audit of the 150 files (\texttt{camera\_aug/provenance\_scan.py}) found no EXIF in any file; all 129 JPEGs use the exact libjpeg quality-75 quantization tables (Pillow's default save) with 4:2:0 chroma, so the last encoder was a generic re-save, and 21 files named \texttt{.jpeg} are PNG.
Gray thumbnails group the files into 76 scenes (coarse correlation $\ge$0.996 within a group, $<$0.3 between groups): 22 singletons, 39 pairs, 12 triples, one quadruple and two quintuples.
Each scene's representative is its highest-resolution real JPEG: 50 are full resolution (4608$\times$3456 or 4352$\times$3264; window side 1838--2032\,px) and 26 exist only at quarter resolution (1152$\times$864; 485--521\,px); the other 74 files are 47 quarter-resolution copies of a representative's exposure, 6 byte-identical duplicates, 10 PNG decodes of a JPEG and 11 notebook-synthetic files (crop, rotation and brightness changes of a real file; training only), so 139 files are real.
Within a scene, a full-resolution file and its 4$\times$ smaller copy correlate at a median 0.86 at fine scale, more than the same image re-encoded at quality 75 (0.76) and far more than a 1-pixel shift ($\approx$0.5): they are one exposure.

\paragraph{Label propagation and the 20-scene subset.}
Every other file of a scene is matched to its representative under the four $90^\circ$ rotations of the representative's corners (normalized cross-correlation at 128\,px per inch), and copies at odd rotations (36 real copies and 9 of the 11 synthetic files) receive the representative's counts with $h$ and $v$ exchanged.
The representative's label always wins (\texttt{annotation\_tool/\allowbreak propagate\_scene\_labels.py}).
The 20 representatives flagged \texttt{iaa} in the released records (stratified by resolution and the frozen counter's weave guess, seed 2027) were drawn under an earlier plan in which only they would be counted twice; they open in blind mode and have no special role.

\paragraph{Re-shoots of one patch.}
A label-free check (\texttt{tools/fabric\_groups\_v2.py}) runs the annotation kit's same-patch matcher (SIFT, RANSAC similarity) on all 2{,}850 pairs of representatives: 2{,}809 pairs give at most 5 inliers and 41 give 16--2{,}504, with none in between, so the kit's pre-set threshold of 12 inliers separates them; all 41 fits have scale 0.98--1.03, and rotations close to within 1$^\circ$ on all 11 triangles of matched scenes.
These 41 pairs of \emph{different} scenes show the same patch of fabric photographed again (other light, shadow, focus or turn).
The scene grouping does not merge them; merging gives the 44 patches (20 single scenes and 24 groups of 2--4 scenes; \texttt{tables/fabric\_groups\_v2.csv}), and the bootstrap clustered by these groups (\texttt{score150.py -{}-groups}) is our patch-level sensitivity analysis.

\paragraph{The controlled resolution pairs.}
The 62 full/quarter pairs (56 distinct full-resolution crops, 47 scenes) sample one exposure at a median 1895 and 475\,px per inch.
\ours's count on the quarter copy differs from the full-resolution count by a median 0.57\% (scene-resampled 95\% CI [0.41, 0.67]; signed median $+0.12$\%), and 9 of 124 counts (7.3\%; scene CI [2.3, 13.3]) move by more than 10\%, mostly octave-type flips (\eg 151$\to$76).
By the quarter copy's px per yarn on the counted axis, the share of such flips is 27\% at 3--4\,px (3 of 11), 8\% at 4--6\,px and 4.5\% at $\ge$6\,px.

\section{Pixels per yarn}
\label{sec:supp_stress}
A phone re-capture simulator (\texttt{camera\_aug/camera\_sim.py}: loupe optics, sensor sampling, mosaic, noise, ISP and JPEG in physical order) brings one full-resolution crop per scene (50) and the 13 legacy crops to a target px per yarn by four routes (\cref{tab:supp_ppy}); its phone profiles are priors anchored on public specifications, so these are simulations, and because the sources are already phone photos, a render can only lose information.
Counts collapse (median change $>$3\%) below 2.5\,px per yarn under ideal resampling or a non-anti-aliased messaging resize, 3.0 behind a Bayer sensor at that pitch and 3.5 through the full budget pipeline, so $\ge$3.5--4\,px per yarn avoids systematic collapse.
Keeping \ours's flips at the $\approx$5\% floor of any resampling needs $\ge$4\,px per yarn for ideal resampling and more than 7 through the phone pipeline: at 7\,px per yarn, a 12.5\,MP photo through the glass ($\approx$1700\,px per inch) resolves about 240\,\TPI, and a standard messaging forward (1280-px long side, $\approx$530\,px per inch) about 75\,\TPI.

\begin{table}[t]
\centering
\footnotesize
\setlength{\tabcolsep}{2.5pt}
\begin{tabular}{@{}lcccc@{}}
\toprule
px per yarn & area & Bayer at pitch & phone at pitch & phone + messaging \\
\midrule
2.0 & 51 / 98 & 54 / 100 & 63 / 100 & 51 / 94 \\
2.5 & \textbf{1.5} / 41 & 29 / 66 & 38 / 74 & \textbf{2.2} / 45 \\
3.0 & 1.3 / 14 & \textbf{2.4} / 44 & 25 / 58 & 1.7 / 20 \\
3.5 & 0.7 / 11 & 1.3 / 29 & \textbf{2.9} / 43 & 1.3 / 17 \\
4.0 & 0.8 / \textbf{6.2} & 1.0 / 19 & 1.2 / 25 & 0.8 / 12 \\
5.0 & 0.4 / 4.5 & 0.8 / 16 & 0.8 / 17 & 0.5 / 9.0 \\
$>$7 & 0.1 / 5.4 & 0.3 / 10 & 0.2 / \textbf{4.6} & 0.2 / 10 \\
\bottomrule
\end{tabular}
\caption{\ours's median absolute count change (\%) / share of counts changing by more than 10\% (\%) against px per yarn on the counted axis, by how the low resolution arises (50 full-resolution scenes and the 13 legacy crops; \texttt{label\_audit\_sweep.csv} of \texttt{camera\_aug/results}). Bold: from this bin on, the median stays $\le$3\% (left) or the flip share stays within 5 points of the resampling plateau (right).}
\label{tab:supp_ppy}
\end{table}

\section{Generative enhancement and hallucinated threads}
\label{sec:supp_genai}
\paragraph{Protocol.}
The protocol was frozen before any restoration model was downloaded or run (README md5 recorded).
Reference sets: R-LEG, the 13 legacy crops with hand counts, and R-HR, the 53 \dataset crops with $S\ge1500$\,px and \ours confidence $\ge$0.574 (the cascade's confidence-only threshold, fitted on synthetic data), scored against \ours's own count on the clean crop (the pre-registered primary pseudo-reference; \cref{tab:supp_genai} and the main paper show each estimator against its own clean count, the pre-registered secondary reference).
R-HR was fixed before the scene audit: its 53 crops are 40 scenes and include copies and notebook-synthetic crops, so its intervals, which resample crops, are optimistic.
Degradations: $\times$4 and $\times$8 \texttt{INTER\_AREA} down-sampling and $\times$4 + JPEG\,$q$50.
Enhancements, all brought back to the reference size: bicubic interpolation (the pre-registered control), SwinIR-M $\times$4~\cite{liang2021swinir}, Real-ESRGAN $\times$4plus~\cite{wang2021realesrgan}, SeedVR2-3B~\cite{wang2026seedvr2} (third-party fp16 repackaging, with its runner's default LAB color correction), InvSR~\cite{yue2025invsr}, FLUX.2 [klein] 4B~\cite{bfl2026flux2klein} and Nano Banana Pro~\cite{google2025nanobananapro} on R-LEG only (4 of 39 requests refused by its content filter, not retried), with one seed (20261008), default settings and every output scored.
Estimators: \ours, the log-Gabor bank and the synthetic CNN; GPT-6.1-sol P1 on R-LEG at $\times$8 only.
The spectral metrics are the share of significant energy at peaks of the counter's whitened spectrum that are neither on the reference lattice nor sampling replicas (``novel''), and the erasure of lattice points.

\paragraph{The control failed (deviation D1).}
After counting only clean, degraded and bicubic crops, we found that bicubic $\times$8 up-sampling makes \ours lock onto the interpolation imaging replica $f_s-f_0$ of the yarn frequency (R-LEG 7.6$\to$301.7\%, R-HR 15.8$\to$239.7\%, with confidence 0.46--0.81), because whitening amplifies weak structure in the empty high band and the lattice prior prefers the finest periodicity; log-Gabor and the CNN are unaffected.
The pre-registered primary hypothesis H1 (no generative restoration lowers \ours's $\times$8 error below bicubic's) was therefore uninformative: every restorer beats the broken control (Holm $p\le0.006$).
Thirteen minutes before any SR or generative output was counted, we added comparisons against counting the degraded crop at native resolution and split the off-lattice energy into sampling replicas and novel peaks; both are deviations and are reported as such.

\paragraph{Results (\cref{tab:supp_genai}).}
Against native-resolution counting, 3 of the 27 comparisons for \ours improve and 3 worsen (95\% bootstrap CIs; 2 and 1 after Holm), the CNN worsens in 22 of 27 (8 after Holm), and SwinIR, the non-generative baseline, never worsens.
Pooled over the three estimators, the 81 generative comparisons give 33 worse, 41 not different and 7 better with the secondary R-HR reference and 21, 45 and 15 with the primary one (the CNN worse in 17 of 27, log-Gabor better in 12 of 27); neither tally was pre-registered.
Post hoc, \ours's legacy differences mostly reflect 2--4 crops: down-sampling cures two clean failures (a6 and r3 wefts), and restorers that resynthesize fine structure bring them back; without a6 and r3, MAPE is 4.2--4.7\% for SwinIR, Real-ESRGAN and Nano Banana Pro against 6.4\% unrestored, and at $\times$8 the within-10\% rate is never worse than unrestored.
SeedVR2 helps \ours on R-HR (3.8\% \vs 15.8\%) by resynthesizing harmonics of fundamentals the low-resolution image still holds, but not on R-LEG (24.1\% \vs 7.6\%, n.s.; legacy a1 read as 229 against its label of 76).
H2 (more off-lattice peaks than bicubic) is not supported: only Real-ESRGAN and InvSR add substantial novel periodicity (on average 13\% and 4--9\% of the significant energy at $\times$8), and Real-ESRGAN exceeds bicubic only marginally ($p=0.055$).
H3 holds for the perceptual metric but not uniformly for counts: every generative output has a mean MUSIQ~\cite{ke2021musiq} at or above that of the real high-resolution reference (significantly in 25 of 27 set--degradation cells), while at $\times$8 PSNR falls to 19--24\,dB for all generative models but SeedVR2: no-reference quality rewards invented detail.
H4: about half of all enhanced crops on which \ours errs by more than 10\% keep a confidence of at least 0.574 (109 of 211), as do unrestored crops (23 of 49), so enhancement does not make silent failures more common; and GPT-6.1-sol's $\times$8 error trends from 11.7\% (bicubic) toward its clean level with SeedVR2 (8.6\%) and Nano Banana Pro (7.9\%, 12 crops), not significantly.
Each main run uses one seed (exploratory: with FLUX.2 a seed change alone moves 21--28\% of \ours's R-HR $\times$8 counts by more than 10\%).

\begin{table*}[t]
\centering
\footnotesize
\resizebox{\textwidth}{!}{\begin{tabular}{llrrrrrr}
\toprule
 & & \multicolumn{3}{c}{MAPE (\%), $\Delta$ vs.\ degraded} & & & \\
Set & Restoration & \ours & log-Gabor & CNN & novel-peak energy & LPIPS$\downarrow$ & MUSIQ$\uparrow$ \\
\midrule
R-LEG & Degraded (LR, no restoration) & 7.6 & 12.7 & 12.4 & 0.0\% & -- & -- \\
R-LEG & Bicubic & 301.7 (+294.1)$^\uparrow$ & 13.6 (+0.9) & 12.2 (-0.2)$^\downarrow$ & 9.9\% & 0.44 & 34 \\
R-LEG & SwinIR-M $\times$4 & 28.2 (+20.6) & 9.0 (-3.6)$^\downarrow$ & 12.3 (-0.1) & 0.0\% & 0.22 & 50 \\
R-LEG & Real-ESRGAN $\times$4plus & 23.4 (+15.9) & 11.5 (-1.2) & 28.1 (+15.7)$^\uparrow$ & 13.3\% & 0.44 & 56 \\
R-LEG & SeedVR2-3B & 24.1 (+16.5) & 9.6 (-3.1) & 21.5 (+9.0)$^\uparrow$ & 2.0\% & 0.23 & 52 \\
R-LEG & InvSR (1 step) & 20.0 (+12.5) & 9.9 (-2.8) & 23.6 (+11.1)$^\uparrow$ & 8.9\% & 0.43 & 60 \\
R-LEG & FLUX.2 klein 4B & 15.0 (+7.4) & 9.8 (-2.8)$^\downarrow$ & 19.0 (+6.6)$^\uparrow$ & 2.3\% & 0.41 & 63 \\
R-LEG & Nano Banana Pro & 5.9 (-2.1)$^\downarrow$ & 8.6 (-2.5)$^\downarrow$ & 18.6 (+5.9)$^\uparrow$ & 1.0\% & 0.28 & 60 \\
\midrule
R-HR & Degraded (LR, no restoration) & 15.8 & 9.5 & 9.5 & 0.0\% & -- & -- \\
R-HR & Bicubic & 239.7 (+223.9)$^\uparrow$ & 8.4 (-1.0) & 9.8 (+0.3)$^\uparrow$ & 5.5\% & 0.48 & 34 \\
R-HR & SwinIR-M $\times$4 & 11.7 (-4.1) & 2.5 (-7.0)$^\downarrow$ & 6.4 (-3.1)$^\downarrow$ & 0.5\% & 0.31 & 45 \\
R-HR & Real-ESRGAN $\times$4plus & 29.1 (+13.3)$^\uparrow$ & 24.8 (+15.4)$^\uparrow$ & 21.5 (+12.0)$^\uparrow$ & 13.4\% & 0.47 & 53 \\
R-HR & SeedVR2-3B & 3.8 (-11.9)$^\downarrow$ & 5.5 (-3.9)$^\downarrow$ & 12.7 (+3.2)$^\uparrow$ & 1.3\% & 0.27 & 50 \\
R-HR & InvSR (1 step) & 19.4 (+3.6) & 16.7 (+7.2) & 14.7 (+5.3)$^\uparrow$ & 4.0\% & 0.39 & 57 \\
R-HR & FLUX.2 klein 4B & 9.9 (-5.9)$^\downarrow$ & 14.0 (+4.6) & 12.0 (+2.5)$^\uparrow$ & 0.7\% & 0.40 & 61 \\
\bottomrule
\end{tabular}
}
\caption{Enhancement before counting at $\times$8 down-sampling. MAPE (\%) and, in parentheses, the paired change against counting the degraded crop at native resolution (a deviation from the pre-registered bicubic control, fixed before any generative output was scored); $^\uparrow$/$^\downarrow$: paired 95\% bootstrap CI excludes 0 (worse/better), uncorrected. R-LEG: hand counts (Nano Banana Pro on the 12 crops it returned, its degraded baseline 8.0\%); R-HR: 53 crops, each estimator against its own clean prediction. Novel-peak energy: mean share of significant spectral energy neither on the reference lattice nor at a sampling replica. LPIPS against the reference; MUSIQ no-reference (clean reference: 47 on R-HR, 52 on R-LEG).}
\label{tab:supp_genai}
\end{table*}

\section{A native-resolution guard against up-sampled input}
\label{sec:supp_guard}
The guard wraps the frozen counter without changing it: per axis it looks in the mean row (column) power spectrum for a \emph{dead band}, an upper frequency range left near the 8-bit quantization level, whose lower edge gives the content cut-off $f_c$ and an effective native side $n_\text{eff}=2f_c$; if there is one, the native side is clearly smaller than the working side and the unguarded count lies outside the band the counter would search at that side, it recounts at side $0.9\,n_\text{eff}$, and otherwise its output is the unguarded estimate, bit for bit (\texttt{methods/guard}).
The idea is post hoc: the failure it targets was first seen on the real sets (deviation D1 of \cref{sec:supp_genai} and the defocus over-count). Only its eight tuned parameters are blind: they were fixed on 2{,}880 fresh synthetic crops against their known native side and exact labels, and frozen (\texttt{guard\_params.json}) before the single real run.
Run once on 1{,}884 real inputs, the guard lowers bicubic $\times$8 error from 301.7 to 6.0\% [2.8, 9.6] on the legacy crops (13 of 13 better; Wilcoxon $p=0.0002$) and from 240.0 to 14.5\% [9.1, 20.5] on R-HR (47 better, 1 worse; scene-resampled $\Delta$ [$-278.7$, $-172.6$]; \cref{sec:supp_genai} reports 239.7\% from its own run of the counter), within 1.6 points of native-resolution counting (7.6\% and 15.8\%; not tested).
At $\times$4 it fires on only 2--8\% of the up-sampled crops, and the gain on R-HR bicubic $\times$4+JPEG ($-6.3$ [$-13.8$, $-0.7$]) is not significant when scenes are resampled ([$-17.1$, $0.0$]).
It fires on none of the 150 clean files (0 of 76 scenes), the 150 crops re-made from the originals, the 66 high-resolution references, the 198 native low-resolution inputs and the 13 clean legacy crops, and on none of the SeedVR2, InvSR, FLUX.2, Nano Banana Pro or Real-ESRGAN outputs (2 of 198 SwinIR outputs): generative restorers fill the empty band, so the guard cannot see their texture.
Of the degraded legacy conditions it was run on (defocus, down-sampling, JPEG, glare and corner error; \texttt{real\_table\_cascade.csv}), it fires only under defocus: at 2--4\,u it removes the over-count (MAPE 186--246\%$\to$14--19\%; at 3\,u over-counts fall from 13 to 1 of 26), but at 6\,u the content cut-off falls below many yarn frequencies, so its recounts leave 23 of 26 counts more than 10\% low (within 10\%: 17$\to$3), as the synthetic set had warned (median error at 6\,u 9.2$\to$59.5\%); over the nine conditions of the cascade (\cref{sec:supp_cascade}, clean included) 9 photo-conditions improve and 14 worsen ($\Delta$MAPE $-22.6$ [$-41.5$, $-5.0$], Wilcoxon $p=0.41$).
It cannot tell up-sampling from blur, as both leave an empty band, so it needs the confidence, or a VLM, behind it.

\begin{table}[t]
\centering
\scriptsize
\setlength{\tabcolsep}{1.5pt}
\resizebox{\ifdim\width>\linewidth\linewidth\else\width\fi}{!}{%
\begin{tabular}{@{}lcccccc@{}}
\toprule
 & & MAPE & \multicolumn{2}{c}{\% of counts} & Both & \\
\cmidrule(lr){4-5}
Method & MAE & [95\% CI] & $\le$5\% & $\le$10\% & $\le$5\% & Weave \\
\midrule
Leave-one-out mean & 19.1 & 28.7 [24.2, 33.6] & 9 & 24 & 0 & -- \\
Log-Gabor bank & 29.7 & 38.3 [33.9, 42.9] & 14 & 18 & 2 & -- \\
\ours (spectral) & 26.1 & 35.1 [30.7, 39.6] & 11 & 20 & 2 & 60 \\
Combined (\ours + log-Gabor) & 25.9 & 34.8 [30.3, 39.2] & 13 & 19 & 2 & -- \\
\ours, ruler crop$^{75/76}$ & 26.9 & 35.9 [31.2, 40.8] & 11 & 20 & 2 & -- \\
Cascade (\ours/GPT) & 25.8 & 34.8 [30.4, 39.5] & 11 & 19 & 2 & -- \\
CNN, synthetic (zero-shot) & 24.4 & 35.2 [30.2, 40.3] & 7 & 15 & 0 & 46 \\
CNN, + pseudo-labels$^\ast$$^{48/76}$ & 22.5 & 29.8 [25.2, 34.5] & 11 & 21 & 1 & 44 \\
\quad without 9 re-shoots (@39)$^{39/76}$ & 21.6 & 29.1 [24.0, 34.7] & 12 & 22 & 1 & 40 \\
GPT-6.1-sol & 19.8 & 26.6 [23.0, 30.5] & 13 & 20 & 5 & 56 \\
Claude Fable 5.1 & 28.7 & 37.1 [33.9, 40.6] & 5 & 9 & 1 & 55 \\
Gemini 3.1 Pro & 21.1 & 31.8 [26.8, 37.2] & 12 & 21 & 3 & 45 \\
Gemini 3.8 Flash & 23.0 & 29.8 [26.2, 33.7] & 10 & 14 & 1 & 46 \\
Qwen3.8-Max-Prime$^{72/76}$ & 23.8 & 29.4 [25.6, 33.1] & 9 & 19 & 2 & 44 \\
Qwen3.8-27B (open) & 58.4 & 75.2 [72.2, 77.8] & 0 & 0 & 0 & 52 \\
Gemma-4-26B (open) & 45.6 & 58.1 [55.4, 60.9] & 0 & 0 & 0 & 48 \\
\midrule
Lattice only$^{73/76}$ & 25.8 & 35.1 [30.3, 39.9] & 13 & 21 & 2 & -- \\
FFT max near-axis peak$^{67/76}$ & 27.7 & 37.4 [32.4, 42.7] & 13 & 21 & 1 & -- \\
FFT tile voting$^{70/76}$ & 30.0 & 44.0 [34.1, 56.3] & 13 & 20 & 0 & -- \\
Deskewed profile + YIN & 31.0 & 40.0 [35.5, 44.5] & 9 & 16 & 2 & -- \\
CountGD (pre-declared) & 285.0 & 404.0 [342.2, 472.6] & 1 & 3 & 0 & -- \\
CountGD, cells $\to$ lattice$^\ddagger$ & 20.8 & 25.6 [21.5, 30.1] & 15 & 27 & 2 & -- \\
Molmo2-8B, P1 & 63.0 & 82.2 [81.0, 83.4] & 0 & 0 & 0 & 51 \\
Molmo2-8B, P4 & 43.3 & 53.7 [49.0, 58.5] & 1 & 4 & 0 & -- \\
Molmo2-4B, P1 & 64.7 & 84.7 [83.6, 85.8] & 0 & 0 & 0 & 58 \\
Molmo2-4B + synthetic LoRA, P1 & 19.4 & 28.2 [24.2, 32.4] & 13 & 23 & 2 & 59 \\
Molmo2-4B + synthetic LoRA, P4s$^\ddagger$ & 15.9 & 23.3 [19.7, 27.4] & 13 & 30 & 1 & -- \\
\texttt{s4-peaks} & 24.0 & 29.1 [25.6, 33.0] & 8 & 16 & 3 & -- \\
CNN 5-fold, training folds' mean$^{\mathrm{m}}$ & 19.7 & 29.5 [25.0, 34.4] & 9 & 18 & 0 & 58 \\
CNN 5-fold FT, ImageNet init$^{\mathrm{m}}$ & 10.8 & 15.6 [13.1, 18.2] & 21 & 43 & 5 & 65 \\
CNN 5-fold FT, synthetic init$^{\mathrm{m}}$ & 10.8 & 15.8 [13.1, 19.0] & 28 & 43 & 8 & 69 \\
\midrule
Octave check$^\ddagger$ & 25.6 & 34.5 [30.0, 39.1] & 12 & 20 & 2 & -- \\
CNN, trained on non-expert$^\ddagger$ & 15.2 & 20.8 [17.3, 25.0] & 12 & 33 & 3 & 54 \\
CNN--VLM ensemble$^\ddagger$ & 14.6 & 19.4 [15.9, 23.3] & 19 & 31 & 7 & -- \\
\bottomrule
\end{tabular}}
\caption{All methods on \dataset against the mean of the two annotators' counts, the reference fixed before labeling (scored per scene, $n=76$). MAE in \TPI; MAPE, $\le$5\%, $\le$10\% and weave in \%; both $\le$5\%: scenes with both counts within 5\% (of those scored); superscripts give coverage when a method did not answer for every scene. Top: the methods of main Tab.~1, every row but the leave-one-out mean fixed before any human count (Tab.~1's other rows, Molmo2-4B + LoRA, the CountGD variant, the CNN trained on the non-expert and the ensemble, are in the middle and bottom blocks; GPT-6.1-sol with the four insets: \cref{tab:supp_prereg}, last block, and Sec.~S6); $^\ast$scored only on the 48 scenes the pseudo-label CNN was not trained on, and (@39) without the 9 of them that re-shoot the fabric patch of a training scene (\cref{sec:supp_cnnaug}). Middle: single classical estimators (\cref{sec:supp_lattice}), the counting specialists and Molmo2-4B (\cref{sec:supp_spec}), \texttt{s4-peaks} (\cref{sec:supp_all150}), and five-fold fine-tuning on the human counts with folds grouped by scene ($^{\mathrm{m}}$trained on this mean, so optimistic; \cref{sec:supp_task}); the six variants with label-exact augmentation or the augmented set (15.5--16.8\%): \cref{sec:supp_task}. $^\ddagger$Post hoc (main Sec.~5): the CountGD variant, the P4s readout, the octave check of \cref{sec:supp_octave}, the CNN fine-tuned on the non-expert's counts with folds grouped by patch, and its per-axis geometric mean with GPT-6.1-sol P1 (the CNN--VLM ensemble). On lines where the annotators differ twofold their mean matches neither reading, so the main paper scores against the textile-trained annotator (A2) instead.}
\label{tab:supp_all150}\label{tab:supp_meanref}
\end{table}

\begin{table*}[t]
\centering
\footnotesize
\setlength{\tabcolsep}{3.2pt}
\resizebox{\ifdim\width>\linewidth\linewidth\else\width\fi}{!}{%
\begin{tabular}{@{}lcccccccc@{}}
\toprule
& \multicolumn{4}{c}{MAPE (\%) [95\% CI] against} & \multicolumn{2}{c}{against A2 (rule 5)} & \multicolumn{2}{c}{against the mean (rule 5)} \\
\cmidrule(lr){2-5}\cmidrule(lr){6-7}\cmidrule(lr){8-9}
Method & A1 & A2 (Tab.~1) & mean (\cref{tab:supp_meanref}) & mean, undisputed (rule 2) & $|\log_2|$ & $>$0.5 (\%) & $|\log_2|$ & $>$0.5 (\%) \\
\midrule
The other annotator & 40.9 [32, 51] & 37.3 [31, 44] & -- & -- & 0.52 & 47 & -- & -- \\
Leave-one-out mean & 41.2 [35, 48] & 33.2 [28, 40] & 28.7 [24, 34] & 25.6 [21, 30] & 0.43 & 32 & 0.38 & 27 \\
\midrule
Log-Gabor bank & 41.6 [37, 46] & 37.7 [32, 44] & 38.3 [34, 43] & 35.0 [29, 41] & 0.77 & 55 & 0.77 & 60 \\
\ours (spectral) & 41.7 [36, 48] & 34.8 [28, 41] & 35.1 [31, 40] & 29.9 [25, 35] & 0.57 & 43 & 0.59 & 50 \\
Combined (\ours + log-Gabor) & 41.4 [36, 47] & 34.4 [28, 41] & 34.8 [30, 39] & 29.6 [24, 35] & 0.57 & 43 & 0.58 & 50 \\
\ours, ruler crop & 41.5 [36, 48]$^{75}$ & 37.9 [31, 45]$^{75}$ & 35.9 [31, 41]$^{75}$ & 29.7 [24, 35] & 0.60 & 45 & 0.59 & 47 \\
Cascade (\ours/GPT) & 41.0 [35, 47] & 34.5 [28, 41] & 34.8 [30, 40] & 28.7 [24, 34] & 0.56 & 46 & 0.57 & 50 \\
\midrule
CNN, synthetic (zero-shot) & 46.2 [38, 54] & 39.3 [32, 48] & 35.2 [30, 40] & 31.3 [27, 36] & 0.47 & 35 & 0.44 & 36 \\
CNN, + pseudo-labels$^\ast$ & 41.6 [35, 50]$^{48}$ & 30.7 [24, 39]$^{48}$ & 29.8 [25, 34]$^{48}$ & 25.5 [20, 32]$^{31}$ & 0.45 & 33 & 0.44 & 40 \\
CNN, 5-fold FT (synthetic init)$^{\mathrm{m}}$ & 30.6 [25, 36] & 22.5 [18, 28] & 15.8 [13, 19] & 11.7 [10, 14] & 0.29 & 16 & 0.22 & 10 \\
CountGD, cells $\to$ lattice$^\ddagger$ & 32.3 [28, 37] & 28.5 [23, 35] & 25.6 [22, 30] & 22.9 [19, 28] & 0.50 & 34 & 0.46 & 31 \\
\midrule
GPT-6.1-sol & 28.8 [25, 33] & 27.1 [22, 33] & 26.6 [23, 30] & 22.8 [19, 27] & 0.45 & 41 & 0.44 & 39 \\
Claude Fable 5.1 & 38.5 [35, 42] & 38.0 [33, 44] & 37.1 [34, 41] & 36.6 [32, 41] & 0.69 & 61 & 0.68 & 66 \\
Gemini 3.1 Pro & 44.3 [37, 53] & 35.9 [30, 42] & 31.8 [27, 37] & 27.9 [22, 34] & 0.46 & 41 & 0.41 & 32 \\
Gemini 3.8 Flash & 33.8 [29, 39] & 31.2 [27, 37] & 29.8 [26, 34] & 25.4 [21, 30] & 0.53 & 43 & 0.50 & 42 \\
Qwen3.8-Max-Prime & 30.3 [27, 34]$^{72}$ & 32.0 [27, 37]$^{72}$ & 29.4 [26, 33]$^{72}$ & 27.0 [23, 32]$^{52}$ & 0.58 & 44 & 0.54 & 46 \\
Qwen3.8-27B (open) & 72.9 [69, 76] & 74.7 [72, 78] & 75.2 [72, 78] & 74.2 [70, 78] & 2.19 & 100 & 2.21 & 100 \\
Gemma-4-26B (open) & 55.9 [53, 59] & 56.7 [53, 60] & 58.1 [55, 61] & 58.2 [55, 61] & 1.32 & 94 & 1.34 & 97 \\
\midrule
Octave check$^\ddagger$ & 41.9 [36, 48] & 33.8 [27, 41] & 34.5 [30, 39] & 29.5 [24, 35] & 0.55 & 40 & 0.57 & 49 \\
CNN, trained on non-expert$^{\ddagger,\mathrm{a}}$ & 28.3 [24, 33] & 28.1 [23, 34] & 20.8 [17, 25] & 17.1 [14, 20] & 0.40 & 34 & 0.31 & 19 \\
CNN--VLM ensemble$^{\ddagger,\mathrm{a}}$ & 25.5 [22, 30] & 25.4 [21, 31] & 19.4 [16, 23] & 16.8 [14, 20] & 0.40 & 29 & 0.31 & 20 \\
\midrule
Molmo2-8B, P1 & 80.7 [79, 82] & 81.7 [80, 83] & 82.2 [81, 83] & 82.3 [81, 84] & 2.55 & 100 & 2.57 & 100 \\
Molmo2-8B, P4 & 53.3 [48, 58] & 52.3 [47, 58] & 53.7 [49, 58] & 50.1 [44, 57] & 1.52 & 78 & 1.53 & 80 \\
CountGD (pre-declared) & 433.0 [365, 509] & 434.4 [357, 520] & 404.0 [342, 473] & 361.9 [288, 451] & 2.01 & 93 & 1.95 & 91 \\
Molmo2-4B + LoRA, P1 (ours) & 47.6 [40, 57] & 25.2 [21, 31] & 28.2 [24, 32] & 24.2 [19, 30] & 0.32 & 22 & 0.38 & 32 \\
Molmo2-4B + LoRA, P4s$^\ddagger$ & 39.4 [32, 48] & 24.1 [19, 30] & 23.3 [20, 27] & 16.6 [14, 20] & 0.31 & 17 & 0.31 & 22 \\
GPT-6.1-sol, four insets & 30.0 [26, 34] & 24.9 [20, 31] & 24.3 [21, 28] & 18.8 [15, 23] & 0.37 & 27 & 0.37 & 33 \\
\bottomrule
\end{tabular}}
\caption{Pre-registered sensitivity analyses (plan for the reference annotator (A2), rules 2, 3 and 5; \cref{sec:supp_task}): the methods of Tab.~1 (Molmo2-4B and the insets: last block) and the five rows trained on human counts or chosen after them (also in \cref{tab:supp_all150}) against the non-expert (A1) alone, the reference annotator (A2) alone, their mean (the reference fixed before labeling; main Tab.~1 uses A2) and the mean without the 72 octave-disputed lines of \cref{tab:supp_disputed} (rule 2: scene MAPE over the remaining axes, 54 scenes, 80 of 152 axes), and the log-ratio endpoint (rule 5): mean $|\log_2(\text{prediction}/\text{reference})|$ over the scene axes and the share of axes off by more than a factor of 1.41 ($|\log_2|>0.5$). Other annotator: one annotator's counts scored against the other (the human noise floor of Sec.~5.1 of the main paper). Superscripts give the number of scenes when a method did not answer every scene. Leave-one-out mean: the mean count of the other scenes of the same reference (rule 2: of the mean). $^{\mathrm{a}}$Trained out of fold on A1's counts, so its A2 column is the blind test. $^{\mathrm{m}}$Trained out of fold on the mean of both annotators (folds grouped by scene). $^\ast$As in Tab.~1. $^\ddagger$Post hoc (main Sec.~5). Last block, added after both annotators' counts were seen and outside the plan (the summary statements of \cref{sec:supp_task} cover only the 20 rows above it): the untuned counting specialists (\cref{sec:supp_spec}; CountGD's readout as declared before the run), Molmo2-4B + LoRA (\cref{sec:supp_lora}) and GPT-6.1-sol with the four magnified insets of \cref{sec:supp_prompts}, all with counts exported before any human count. 95\% CIs: scene bootstrap, as in Tab.~1. Scenes with both counts within 5\%: at most 7 of 76 against A1 (GPT-6.1-sol) and 8 against A2 (Combined, GPT-6.1-sol). Predicting each scene's reference count from the patch's other photos gives a MAPE (\%) of \FNrowHumMape{} (\FNhumN{} scenes; post hoc), a property of the reference; each annotator's re-shoot repeatability is in Tab.~2 of the main paper and \cref{sec:supp_iaa}.}
\label{tab:supp_prereg}
\end{table*}

\begin{table*}[t]
\centering
\scriptsize
\setlength{\tabcolsep}{3pt}
\begin{tabular}{@{}rcrrrrr@{\hspace{14pt}}rcrrrrr@{}}
\toprule
Scene & ax & A1 & A2 & A2/A1 & \ours & GPT & Scene & ax & A1 & A2 & A2/A1 & \ours & GPT \\
\midrule
1 & $h$ & 91 & 44 & 0.48 & 45 & 44 & 55 & $h$ & 52 & 156 & 2.99 & 26 & 52 \\
3 & $h$ & 101 & 60 & 0.59 & 62 & 96 & 55 & $v$ & 41 & 123 & 3.00 & 20 & 41 \\
3 & $v$ & 124 & 68 & 0.55 & 70 & 72 & 56 & $h$ & 54 & 159 & 2.94 & 27 & 52 \\
4 & $h$ & 132 & 64 & 0.48 & 64 & 62 & 56 & $v$ & 41 & 121 & 2.95 & 21 & 40 \\
4 & $v$ & 135 & 67 & 0.50 & 129 & 70 & 59 & $h$ & 106 & 59 & 0.55 & 120 & 60 \\
5 & $v$ & 99 & 49 & 0.49 & 51 & 50 & 59 & $v$ & 54 & 79 & 1.46 & 19 & 40 \\
6 & $h$ & 140 & 64 & 0.46 & 64 & 62 & 60 & $h$ & 49 & 112 & 2.29 & 105 & 52 \\
6 & $v$ & 126 & 65 & 0.52 & 129 & 70 & 60 & $v$ & 58 & 136 & 2.34 & 143 & 72 \\
7 & $h$ & 27 & 41 & 1.52 & 37 & 28 & 61 & $h$ & 51 & 113 & 2.21 & 105 & 104 \\
12 & $v$ & 42 & 81 & 1.93 & 61 & 50 & 61 & $v$ & 58 & 125 & 2.15 & 35 & 74 \\
14 & $h$ & 39 & 74 & 1.90 & 75 & 48 & 62 & $v$ & 58 & 84 & 1.45 & 40 & 42 \\
14 & $v$ & 59 & 99 & 1.68 & 116 & 56 & 64 & $h$ & 37 & 64 & 1.73 & 73 & 34 \\
15 & $h$ & 51 & 103 & 2.02 & 105 & 52 & 65 & $h$ & 62 & 99 & 1.60 & 103 & 96 \\
15 & $v$ & 70 & 113 & 1.61 & 35 & 76 & 76 & $h$ & 108 & 62 & 0.57 & 64 & 62 \\
17 & $v$ & 147 & 82 & 0.56 & 151 & 80 & 76 & $v$ & 107 & 54 & 0.50 & 107 & 66 \\
18 & $h$ & 178 & 70 & 0.39 & 87 & 90 & 84 & $v$ & 73 & 38 & 0.52 & 40 & 40 \\
18 & $v$ & 144 & 77 & 0.53 & 162 & 136 & 86 & $v$ & 31 & 53 & 1.71 & 54 & 52 \\
20 & $h$ & 43 & 68 & 1.58 & 84 & 38 & 87 & $h$ & 40 & 64 & 1.60 & 75 & 56 \\
20 & $v$ & 27 & 54 & 2.00 & 53 & 52 & 89 & $v$ & 28 & 52 & 1.86 & 52 & 46 \\
24 & $h$ & 29 & 71 & 2.47 & 74 & 40 & 91 & $v$ & 92 & 63 & 0.68 & 71 & 60 \\
24 & $v$ & 87 & 45 & 0.52 & 160 & 38 & 92 & $h$ & 118 & 62 & 0.53 & 63 & 60 \\
25 & $h$ & 85 & 51 & 0.60 & 110 & 80 & 92 & $v$ & 107 & 58 & 0.54 & 108 & 72 \\
25 & $v$ & 76 & 39 & 0.51 & 39 & 54 & 93 & $h$ & 114 & 53 & 0.46 & 55 & 60 \\
28 & $h$ & 72 & 137 & 1.91 & 86 & 90 & 95 & $h$ & 70 & 136 & 1.94 & 86 & 72 \\
28 & $v$ & 129 & 83 & 0.64 & 149 & 68 & 98 & $v$ & 35 & 53 & 1.51 & 32 & 32 \\
31 & $h$ & 39 & 23 & 0.59 & 67 & 64 & 102 & $h$ & 40 & 120 & 3.00 & 20 & 40 \\
31 & $v$ & 30 & 42 & 1.42 & 54 & 80 & 102 & $v$ & 53 & 152 & 2.87 & 26 & 52 \\
32 & $h$ & 28 & 49 & 1.75 & 52 & 46 & 105 & $h$ & 102 & 51 & 0.50 & 52 & 60 \\
32 & $v$ & 43 & 62 & 1.44 & 60 & 68 & 105 & $v$ & 70 & 36 & 0.51 & 36 & 34 \\
35 & $h$ & 129 & 70 & 0.54 & 36 & 72 & 106 & $h$ & 37 & 17 & 0.46 & 55 & 56 \\
38 & $v$ & 93 & 135 & 1.45 & 142 & 76 & 107 & $h$ & 93 & 53 & 0.57 & 60 & 58 \\
39 & $h$ & 111 & 56 & 0.50 & 58 & 56 & 108 & $h$ & 112 & 57 & 0.51 & 58 & 56 \\
43 & $v$ & 83 & 56 & 0.67 & 56 & 46 & 124 & $h$ & 46 & 70 & 1.52 & 36 & 35 \\
45 & $v$ & 81 & 43 & 0.53 & 43 & 48 & 124 & $v$ & 44 & 63 & 1.43 & 32 & 32 \\
48 & $h$ & 84 & 44 & 0.52 & 45 & 44 & 126 & $h$ & 31 & 76 & 2.45 & 74 & 48 \\
50 & $v$ & 58 & 116 & 2.00 & 70 & 120 & 128 & $h$ & 33 & 65 & 1.97 & 34 & 36 \\
\bottomrule
\end{tabular}
\caption{The 72 lines (50 scenes) on which the two annotators differ by more than a factor of 1.41 ($|\log_2(\mathrm{A2}/\mathrm{A1})|>0.5$; rule fixed before the reference annotator's counts were opened): both annotators' counts (threads per window side), their ratio, and \ours's and GPT-6.1-sol P1's counts. On these lines the reference does not fix the octave.}
\label{tab:supp_disputed}
\end{table*}

\section{Pseudo-labels for the CNN}
\label{sec:supp_cnnaug}
Label-exact augmentation applies optics, phone re-capture and geometric transforms whose effect on the yarn count is known in closed form (\texttt{cnn/augment\_optics.py}; verification: extended supplement); augmented copies of the photos: \cref{sec:supp_task}.

\paragraph{Pseudo-label training (pre-declared).}
The scene-split protocol (\texttt{cnn/runs\_aug/PREREG.json}, written before training) accepts a scene if \ours's confidence on its representative is at least the median over the 76 representatives (0.848) and the fused count comes from the lattice: 38 scenes, of which 10 are held out (seed 20261008) and 28 train.
Variants V1 (synthetic only), V2 (with label-exact augmentation), V3 (real pseudo-labels only), V4 (synthetic pretraining, then pseudo-label fine-tuning) and V5 (V4 with test-time augmentation) were each trained for ResNet-18 and EfficientNet-B0; every choice used the pre-declared score (synthetic-validation MAPE and held-out pseudo-label MAPE), and the legacy crops were predicted once per final variant, after selection; a self-check found no scene shared between training and held-out sets and no legacy crop among the 150 (12 checks passed).
EfficientNet-B0 V4 is the pseudo-label row of Tab.~1 (with V5, its test-time-augmented copy, it has the best pre-declared selection score, 23.7, though three seeds of one recipe span 6 points of this score; \texttt{cnn/runs\_aug/selection.json}), added to the tables after every final variant's legacy score was known: on the legacy crops it reaches 10.7\% [7.0, 14.9], $-5.3$ points [$-9.3$, $-1.7$] against the synthetic-only CNN (bootstrap, Holm 0.018 over ten variants; per-count and per-image Wilcoxon not significant after Holm, $p=0.13$--0.44), level with \ours ($-0.9$ [$-10.2$, $+6.3$]).
On \dataset it is scored only on the 48 scenes it was not trained on; nine of them re-shoot a training scene's patch (8, 9, 15, 38, 61 and 103, and 45, 52 and 110 of the ten that selected the variant), and without them it scores 29.1\% against the mean (\cref{tab:supp_all150}; 29.8\% on 48) and, post hoc, 31.7\% against the reference (30.7\% on 48).
The nine other final variants reach 26.3--33.9\% against the mean, six of them below its 29.8\% (V1 and V2 on all 76 scenes, V3--V5 on the same 48), and, post hoc, 27.2--36.4\% against the reference, where four of the 48-scene variants score below its 30.7\%.
A pre-registered one-family-out augmentation ablation and the student-teacher comparison are in the extended supplement.

\section{Half-or-double readings on \dataset}
\label{sec:supp_octave}
\paragraph{Against the reference (post hoc).}
On the band of main Sec.~5.2, \ours's \BToursOctLinesNG{} half-or-double lines are 24 halvings and 12 doublings, GPT-6.1-sol P1's 37 are 34 halvings and 3 doublings, and the non-expert's \BOlinesBigOct{} are 30 doublings and 18 halvings; choosing \ours's octave with the reference counts, the upper bound of main Sec.~5.2, changes 65 of the 152 lines, as it also corrects threefold and larger misses.

\paragraph{Spectral diagnostic (post hoc).}
Against the non-expert, with the stricter band $\bigl||\log_2 r|-1\bigr|\le0.14$ and computed before the reference counts were delivered: on its \FNaTwoHalfN{} halved axes the counter's chosen peak is the on-axis point at half the non-expert's frequency on \FNaTwoHalfOnAxis, and the plain-weave point $(\tfrac12,\tfrac12)$, about 45$^\circ$ off the axis and outside the $\pm15^\circ$ search cone, on none; a peak at the non-expert's frequency is among the lattice's candidates on \FNaTwoHalfCand{} but weaker, and the score books it as the chosen peak's second harmonic.
On the \FNaTwoDoubleN{} doubled axes a harmonic of that frequency wins on only \FNaTwoDoubleHarm, and \FNaTwoDoubleNoPeak{} show no spectral peak (whitened $z<3$) there at all.

\paragraph{Rule 4.}
On the \BNrfourN{} lines (band 0.3) where the counter reads about half or double the non-expert (A1), the reference annotator (A2) is within 10\% of A1 on \BNrfourNGSA{} and of the counter on \BNrfourNGours{} (disputed lines: \cref{tab:supp_disputed}), the third outcome of the rule; the rule's written count, 59 lines in 47 scenes, matches no band (bands of 0.01 to 0.79 give 8 to 113 lines), and every band gives this outcome.

\paragraph{Octave check (octave-v1).}
For each axis the check shears the pre-processed crop so that the counted yarns are axis-parallel, averages it along the yarns in eight bands, and resynthesizes one mean period from the first three harmonics of the counted period, phase-aligned over windows of six periods.
If this mean period shows a second ridge 0.35 to 0.5 periods from the main ridge with at least 15\% of its prominence, the period holds two yarns and the count is doubled (the test is repeated once at the new count, and never doubles beyond the analyzed band); it never halves a count, because a missing second ridge is also what alternately colored yarns look like.
Its threshold was chosen on 750 synthetic renders, the 13 legacy crops and label-free agreement between re-shoots, resolution copies and rotated copies of the 150 photos, and frozen before the reference annotator's counts (post hoc on the non-expert's).
It changes \FNoctAxesChanged{} of 152 scene axes and \ours's MAPE against the mean of the two annotators by \FNoctFirstDiff{} points \FNoctFirstDiffCI{} (post hoc).

\paragraph{Learned selector (octsel-v1, post hoc on the non-expert's counts).}
It multiplies each of the counter's counts by $2^k$, $k\in\{-1,0,+1\}$, with a class-balanced multinomial logistic regression on twelve label-free features (log ratios of the counter's stage estimates, of its alternative count and of GPT-6.1-sol's count to its final count; the whitened evidence at the counted frequency, its double and its half; the confidence; px per yarn).
Its regularization is chosen by nested cross-validation grouped by patch, and it is trained on the non-expert's counts.
Out of fold it changes \ours's MAPE by $-4.3$ points [$-10.0$, $+1.3$] against the non-expert (A1) and by \FNoctselMinusOurs{} \FNoctselMinusOursCI{} against the mean of the two, and under every reference it stays above GPT-6.1-sol P1 (by 8.7, 7.0 and 6.0 points against A1, the mean and A2).
Its decision rule was changed once after the first out-of-fold results; both rules were frozen before the reference annotator's counts. The non-expert's counts repeat less well across re-shoots than the reference annotator's (\cref{sec:supp_iaa}), so the null may reflect label noise as well as missing information.

\paragraph{Blind tests (pre-registered records, against A2).}
The records of both checks, frozen before the reference annotator's counts (\cref{sec:supp_task}), name those counts as their primary reference (\texttt{methods/octave/eval\_octave\_v1.py}).
There octave-v1 and octsel-v1 have a MAPE of 33.8 and 33.2\%, against 34.8\% for \ours and 27.1\% for GPT-6.1-sol P1, with both counts within 5\% on 7.9, 11.8, 7.9 and 10.5\% of the scenes; they trail GPT-6.1-sol P1 by $+6.7$ points [$+1.0$, $+12.7$] and $+6.0$ [$+1.6$, $+10.6$], and differ from \ours by \FNoctBlindDiff{} \FNoctBlindDiffCI{} (octave-v1) and $-1.7$ [$-7.3$, $+3.7$] (octsel-v1, which picks the right $k$ on 58.6\% of the axes against 57.2\% for always keeping the count).
In the secondary tests of its record, octsel-v1 trails the CNN--VLM ensemble by $+7.8$ points [$+3.5$, $+12.1$], the first-specified argmax rule reaches 46.8\% and the all-data model (fitted on all 76 scenes of the non-expert (A1)) 28.7\% ($-6.1$ [$-11.2$, $-1.3$] \vs \ours, $+1.5$ [$-2.0$, $+5.1$] \vs GPT-6.1-sol P1); its exploratory geometric means with GPT-6.1-sol P1 and with the ensemble reach 29.0 and 27.8\%, above GPT-6.1-sol P1 (27.1\%) and the ensemble (25.4\%) alone.

\section{All methods on \dataset}
\label{sec:supp_all150}
\Cref{tab:supp_all150} lists the scored methods against the mean of the two annotators' counts, scored per scene ($n=76$); the non-selected CNN variants are summarized in \cref{sec:supp_task,sec:supp_cnnaug}. Per-file scores over the 139 real files of the labeled scenes (scene-clustered CIs; \texttt{tables/main150\_perfile.csv}) differ from the per-scene MAPE by at most 2.5 points for the ten methods scored (\ours 34.9 \vs 35.1\%, GPT-6.1-sol P1 26.2 \vs 26.6\%, Gemini 3.8 Flash 27.3 \vs 29.8\%).

\texttt{s4-peaks} is a training-free baseline from the scaffold study of \cref{sec:supp_prompts}: the peak count of the brightness profiles of a central horizontal and a central vertical band (\texttt{scipy} \texttt{find\_peaks}, prominence 0.5, set on synthetic images); its counts of the 150 photos were exported before any human count existed (\texttt{vlm/\allowbreak scaffold/\allowbreak results/\allowbreak rect150\_S3\_\allowbreak predictions\_wide.csv}).
Against the mean it reaches 29.1\% [25.6, 33.0], $-6.0$ points [$-9.9$, $-2.0$] \vs \ours (Wilcoxon $p=0.004$), and is level with GPT-6.1-sol P1 ($+2.5$ [$-1.6$, $+6.4$]) and the leave-one-out mean ($+0.4$ [$-5.7$, $+6.1$]); against the reference annotator (A2; blind, as the unmarked rows of main Tab.~1, with post-hoc paired tests as in main Sec.~5.1) it reaches 27.9\% [23.3, 33.0], $-7.0$ [$-12.8$, $-1.3$] \vs \ours (Wilcoxon $p=0.022$), $+0.7$ [$-3.7$, $+5.2$] \vs GPT-6.1-sol P1 and $-5.3$ [$-12.0$, $+0.9$] \vs the leave-one-out mean.

\texttt{combined-v1} (main Sec.~4.2) takes the geometric mean of the two counts where the natural log of their ratio is at most 0.18 in magnitude; on its pre-registered held-out synthetic test (\BOcombSynN{} counts) it raised the share of counts within 5\% from \BOcombSynOurs{} to \BOcombSynComb\% (paired 95\% CI \BOcombSynCI{} points). Its pre-registered tests on the human counts (run because that test passed) give $-0.3$ points [$-0.5$, $-0.1$] \vs \ours (Wilcoxon $p=0.055$; 34 scenes closer, 35 farther) and $-3.5$ [$-7.2$, $+0.1$] \vs the log-Gabor bank ($p=0.23$), against the mean of the two annotators.

\FloatBarrier
{
    \small
    \bibliographystyle{ieeenat_fullname}
    \bibliography{paper_refs}
}